\documentclass[journal]{IEEEtran}
\ifCLASSINFOpdf
  \usepackage[pdftex]{graphicx}
  \DeclareGraphicsExtensions{.jpg,.pdf,.jpeg,.png}
\else
\fi
\usepackage{multirow}
\usepackage{makecell}

\usepackage{amsmath}
\usepackage{algorithm}
\usepackage{algpseudocode}
\usepackage{amsmath}
\usepackage{graphics}
\usepackage{epsfig}
\ifCLASSOPTIONcompsoc
 \usepackage[caption=false,font=normalsize,labelfont=sf,textfont=sf]{subfig}
\else
 \usepackage[caption=false,font=footnotesize]{subfig}
\fi
\usepackage{graphicx}

\begin{document}
\title{Brain Inspired Cognitive Model with Attention \\for Self-Driving Cars}

\author{Shitao~Chen, Songyi~Zhang, Jinghao~Shang, Badong~Chen, Nanning~Zheng*,~\IEEEmembership{Fellow,~IEEE}%
\thanks{S. Chen, S. Zhang, J. Shang, B. Chen and N. Zheng are with the Department of Electronic and Information Engineering, Xi'an Jiaotong University, Xi'an, Shaanxi 710049, P.R. China. E-mail:  {chenshitao, zhangsongyi}@stu.xjtu.edu.cn; {chenbd, nnzheng}@mail.xjtu.edu.cn.}
\thanks{* Correspondence: nnzheng@mail.xjtu.edu.cn}
}


\markboth{}%
{Shell \MakeLowercase{\textit{et al.}}: Bare Demo of IEEEtran.cls for IEEE Journals}



\maketitle

\begin{abstract}
% There are two major vision-based frameworks for self-driving cars. One is perception-driven approach which uses precise perception results to plan and control vehicles. The other is an end-to-end system that directly maps input images to driving behaviors. In this paper, we define the cognitive map for traffic scenes and present a cognitive map based attention model (CMAM) which combines the advantages of the two methods above. Our model is inspired by the cognition and attention mechanism in brain and built for solving three self-driving tasks jointly: \romannumeral1) detection of the free space and boundaries of the current and adjacent lanes.  \romannumeral2)estimation of obstacle distance and vehicle attitude, and \romannumeral3) learning of driving behavior and decision making from human driver. More significantly, the proposed model could accept external navigating instructions during an end-to-end driving process. For evaluation, we construct a large-scale road-vehicle dataset which contains more than 300000 different road images captured by three cameras in our self-driving car. Simultaneously, human driving activities and vehicle states are recorded. The experiment results on this dataset and the simulation environment demonstrate the effectiveness of our method.
Perception-driven approach and end-to-end system are two major vision-based frameworks for self-driving cars.
However, it is difficult to introduce attention and historical information of autonomous driving process, which are the essential factors for achieving human-like driving into these two methods. In this paper, we propose a novel model for self-driving cars named brain-inspired cognitive model with attention (CMA). This model consists of three parts: a convolutional neural network for simulating human visual cortex, a cognitive map built to describe relationships between objects in complex traffic scene and a recurrent neural network that combines with the real-time updated cognitive map to implement attention mechanism and long-short term memory. The benefit of our model is that can accurately solve three tasks simultaneously:
\romannumeral1) detection of the free space and boundaries of the current and adjacent lanes. 
\romannumeral2)estimation of obstacle distance and vehicle attitude, and 
\romannumeral3) learning of driving behavior and decision making from human driver. 
More significantly, the proposed model could accept external navigating instructions during an end-to-end driving process. For evaluation, we build a large-scale road-vehicle dataset which contains more than forty thousand labeled road images captured by three cameras on our self-driving car. Moreover, human driving activities and vehicle states are recorded in the meanwhile. 

\end{abstract}


% \fi
\IEEEpeerreviewmaketitle
\begin{figure*}[ht]
 \centering
 \includegraphics[width=7.0in]{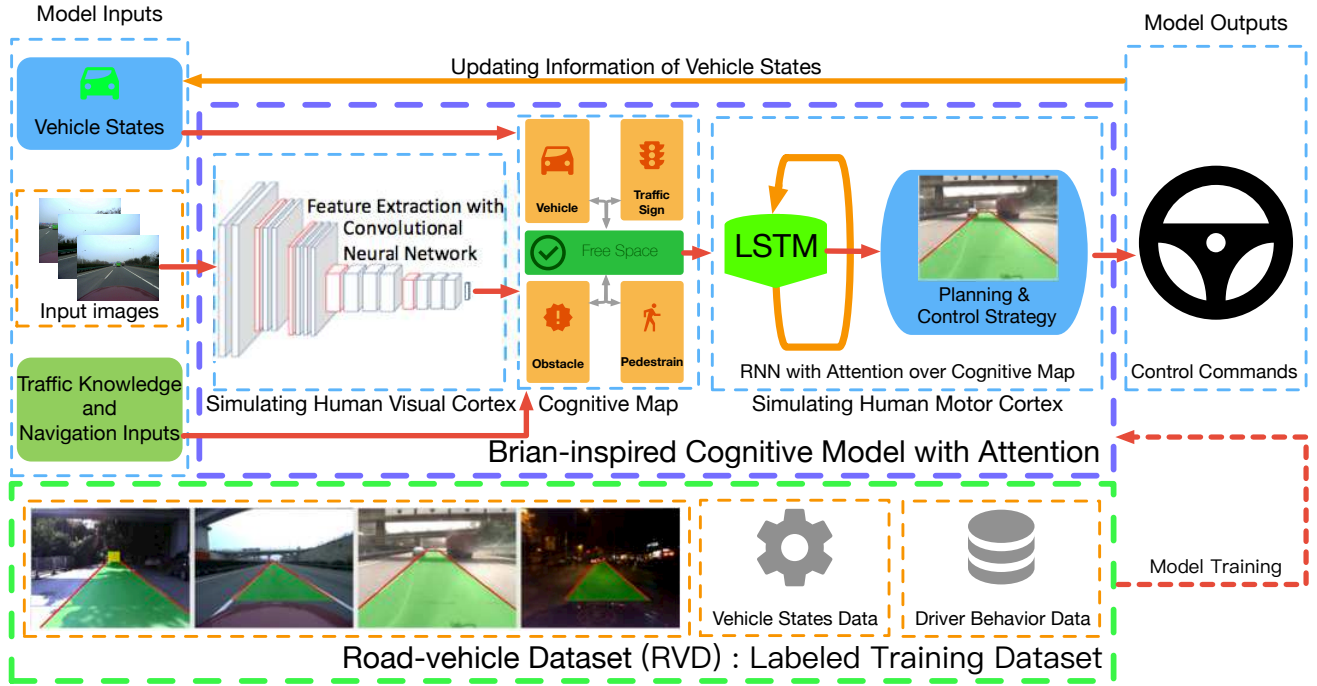}
 \caption{Framework of our cognitive model with attention (CMA). The road images are first processed by multiple convolutional neural networks to simulate the function of human visual cortex. A cognitive map consists of vehicle states, navigation inputs and perception results. Based on the information of cognitive map, recurrent neural network models the attention mechanism by historical states and scene data in time to form driving decision in each computing step.}
 \label{csttt}
\end{figure*}

\section{Introduction}
% \IEEEPARstart{A}{utomatically}
% scene understanding is the core technology of self-driving cars,as well as one of the primary goals of computer vision.
% It is well known that most of related information required for self-driving cars can be obtained through vision sensors.
% During the recent decades, considerable progress and development have been achieved in the study of vision-based self-driving cars.  Similarly, humans can only rely on vision to drive a car. Besides, the mechanism of attention can help people choose the effective data in memory to determine what objects exist in the current image, but also to explore the relationship between these objects to form correct decisions in current moment.
% Therefore, it is a significant research to develop a self-driving car only uses cameras \cite{xue2017vision,tsugawa1994vision,turk1988vits} and explore the implementation of attention mechanism in self-driving cars. 

\IEEEPARstart{A}{utomatically}
scene understanding is the core technology for self-driving cars, as well as a primary pursuit of the computer vision.
During the recent decades, considerable progress and development have been achieved in the field of vision-based self-driving cars.
It is well known that most of related information required for self-driving cars can be obtained by cameras, which is originally inspired by humans' driving behaviors. Besides, the mechanism of attention can help people choose the effective data in memory to determine the objects existing in the current image and their relationships, so as to form correct decisions in current moment.
Therefore, it is significant to develop a self-driving car only based on vision \cite{xue2017vision,tsugawa1994vision,turk1988vits}, where the mechanism of attention is felicitously implemented.

Nowadays, there are two popular vision-based paradigms for self-driving cars: the perception-driven method and the end-to-end method. For the \textbf{perception-driven method} \cite{leonard2008perception}, it is required to establish a detailed representation of the real world. By fusing multi-sensor data, a typical world representation usually contains descriptions of the prime objects in various traffic scene, including (but not limited to) lane boundaries, free space, pedestrians, cars, traffic lights and traffic signs. With such descriptions, the path planning module and the control module are used to determine the actual movement of the vehicle. 
% In the path planning process, it is often necessary to manually design a suitable motion trajectory with some predetermined rules, relying on accurate perception results and high-precision map information. 
In the path planning process, besides accurate perception results and high-precision map information, it is often necessary to design some assistant rules manually.
Such a motion trajectory needs to be adjusted and updated in real-time according to the state of the vehicle at each moment, taking the temporal dependences into account, so as to form a correct trajectory sequence. 
With the motion trajectory calculated by planning module, the vehicle is steered to track each task point on the planned path under the guidance of high-precision positioning information.

As for the \textbf{end-to-end method} \cite{bojarski2016end}, based on the breakthrough of convolutional neural networks (CNN) \cite{krizhevsky2012imagenet} and GPU technology, a deep neural network is able to learn the entire processing pipeline needed for controlling a vehicle, direct from the human driving behaviors. Instead of using hand-crafted features as in perception-driven method, we enable the CNN to learn the most valuable image features automatically and directly map them to the control of the steering angle. Since the actual control of the car only relates to the velocity and steering angle, such method of directly mapping images to the control of direction, is more efficient and effective in some scenarios.

The perception-based method was the most widely used one in the past decades. It can be applied to most challenging tasks, but the disadvantage is that all features and task plans are manually designed, and the entire system lacks the self-learning ability. In recent years, the end-to-end learning strategy for self-driving gradually boomed with the success of deep learning \cite{lecun2015deep}. End-to-end strategy merely requires some visual information, and is capable of learning from human driving behaviors. However, the disadvantage is, when the system structure is simple, the external information is unable to be introduced in to control the behavior of the self-driving system. Therefore, while the system is running, we have no way to know where the vehicle is going, neither can we control the system as well. In the meanwhile, temporal information has never been considered in this end-to-end process.
% For the reasons above, this kind of system  can barely do well in some simulation games or demonstration work, but is not suitable for a wide range of applications in real-time traffic environment.

In our point of view, it is believed that a highly effective and reasonable autonomous system should be inspired by the cognitive process of human brain. First of all, it is able to perceive the environment as rationally as the visual cortex, and then to process the perception results in a proper way. After that, the system plays a role of the motor cortex to plan and control the driving behaviors. And in the whole process, the concept of human-computer collaborative hybrid-augmented intelligence \cite{Zheng2017hunhe} is well referred, so that the self-driving system can learn smartly from human driving behaviors.

In this paper, we aim to build a brain-inspired cognitive model with attention. When a person view a scene, the message flows through LGN to V1, onward to V2, then to V4 and IT \cite{goodfellow2016deep}, which occurs within the first 100ms of a glance to objects. This process is proved to be highly similar to the operating principle of the convolutional neural network. Thus in our model, we adhere to apply CNNs for the processing of the visual information, which is a simulation of the visual cortex to process information. Similarly, as in \emph{On Intelligence}, Jeff Hawkins argues \cite{hawkins2007intelligence} that time holds the vital place in brain when solving a problem. We believe that brain has to deal with spatial and temporal information simultaneously, since spatial patterns need to be coincident with temporal patterns. Therefore, we need to simulate the functions of motor cortex, which means, in dealing with planning and control problems, a long-term memory must be considered to form the optimal driving strategy for the current.
With this motivation, it is necessary to introduce the attention mechanism into the cognitive computing model for self-driving cars, which allows the model to choose reasonable information from a large set of long-term memory data at each computing step.
% we want to create a temporal model, where the perception results are coincident with the temporal patterns. 

% planning and control decisions must be made with a long-term information, so as .  
Moreover, Mountcastle \emph{et al.} \cite{Mountcastle1978} points out that the functional areas in the cerebral cortex have similarities and consistency. He believes the regions of cortex that control muscles are similar to the regions which handle auditory or optical inputs in structure and function. Inspired by this, we argue that the recurrent neural network (RNN), which performs well in processing sequential data and has been successfully applied in video sequence classification and natural language processing tasks, is also capable to solve planning and control problems simultaneously as human motor cortex. The discussion above is an important motivation for us to implement planning and control decision with RNN.

In order to introduce attention mechanism into the proposed cognitive model, and to solve the problem that general end-to-end models cannot introduce external information to guide, we define the concept of cognitive map in real traffic. The term of \textbf{cognitive map} was first coined by Edward Tolman \cite{tolman1948cognitive} as a type of mental representation of the layout of one's physical environment. Thereafter, this concept was widely used in the fields of neuroscience \cite{mcnaughton2006path} and psychology. The research results on cognitive map in these areas provide an important inspiration for us to construct a new model of autonomous driving. To apply this concept to the field of self-driving, combining with our work, \textbf{cognitive map for traffic scene} is built to describe the relationship between objects in complex traffic scene.
%认知地图都是在Neuroscience和psychology中定义的，所以我们第一次定义认知地图在real traffic%
%之前关于认知地图的研究都在心里和神经科学当中，所以我们首次针对无人驾驶任务，给出认知地图in real traffic的定义
%A cognitive map is a mental picture or image of the layout of one's physical environment. The term was first coined by a psychologist named Edward Tolman in the 1940s. 
%Cognitive maps are mental representations of physical locations. 
It is a comprehensive representation of the local traffic scene, including lane boundary, free space, pedestrian, automobile, traffic lights and other objects, as well as the relationships between them, such as direction, distance, etc. Furthermore, the prior knowledge of traffic rules and the temporal information are also taken into consideration. The cognitive map defined in this paper is essentially a structured description of vehicle state and scene data in the past. This description forms the memory of a longer period of time. The proposed cognitive model, in which the cognitive map combines long-short term memory, mimics the human driving ability to understand about traffic scene and to make decisions of driving.

In precise, our framework first extracts valid information from the traffic scene of each moment by a convolutional neural network to form the cognitive map, which contains the temporal information and a long-term memory. On this basis, we add external control information to some descriptions of the cognitive map, e.g., the guidance information from the navigation map. And finally, we utilize a recurrent neural network to model attention mechanism based on historical states and scene data in time, so as to perform path planning and control to the vehicle.

% With all these above, in our cognitive model, a novel end-to-end self-driving framework with attention inspired by human brain has come to form. We name this framework as cognitive map based attention model (CMAM). It is able to handle the spatial and temporal relationships respectively, so as to perform the basic self-driving missions. In this paper, we achieve the target to enable the self-driving system to perceive only by vision sensors, as well as to perform planning and control to the vehicle with the attention mechanism, without the help of position sensor such as GPS. 
With all above, in our model, a novel self-driving framework combined with attention mechanism has come to form, which is inspired by human brain. The framework is named brain-inspired cognitive model with attention (CMA). It is able to handle the spatial-temporal relationships, so as to implement the basic self-driving missions. In this paper, we realized a self-driving system with only vision sensors. It performs well in making path planning and producing control commands for vehicles with attention mechanism.
Fig. \ref{csttt} shows the main scheme of our CMA method. The remainder of the paper is organized as follows: in section \uppercase\expandafter{\romannumeral2}, we review some previous studies of self-driving cars; in section \uppercase\expandafter{\romannumeral3}, we describe our approach in detail; in section \uppercase\expandafter{\romannumeral4} and \uppercase\expandafter{\romannumeral5}, we present a large-scale labeled self-driving dataset and the evaluation of our method; finally section \uppercase\expandafter{\romannumeral6} concludes the work.

\section{Related Work}
% \section{Related Work and Our contributions}

% In this section, we review some previous works related to our paper. Furthermore, the main contributions of our study are referred in detail, comparing with other researches in the literature.
% \subsection{Related Work}
In the past decades, remarkable achievements \cite{chen2015deepdriving,xue2017vision,leonard2008perception} have been reached with  perception-driven method in the filed of self-driving cars. Several detection methods for car and lane boundary have been proposed to build a description of the local environment.

Many lane detection methods in \cite{wang2004lane,assidiq2008real,li2014multiple,he2010lane} have been developed to locate the lane position with canny edge detection or hough transformation. The defect of these methods is that they lack some geometric constraints to locate the arbitrary lane boundary. Therefore, Nan \emph{et al.} \cite{nan2016efficient} presented a spatial-temporal knowledge model to fit the line segments, which finally outputs the lane boundaries. Huval \emph{et al.} \cite{huval2015empirical} introduced a deep learning model to perform lane detection at a high frame rate. Different from the traditional lane boundary detection approach whose output is the pixel location of the lane boundary, the work \cite{chen2015deepdriving} represented a novel idea which uses convolutional neural network to map an input image directly to a deviation between vehicle and lane boundary. With this method, the output of the neural network can be directly used in controlling the vehicle, without coordinate transformation. 
The limitation of this model is that per-training for a specific vehicle is needed. 
% But the limitation is that each model of this kind is trained for a specific vehicle and it can not be applied to all self-driving cars.

For object detection task, researches \cite{girshick2016region,redmon2016you} adopt the method of generating a bounding box to describe the location of the object. However, in the self-driving task, it is not necessary to get a precise location of the bounding box. We only need to know if there is a obstacle in our lane and how far the obstacle is. Thus it is a more convenient and efficient way to represent the obstacle as a point instead of a bounding box.

The concept of end-to-end learning method was originally inspired by Pomerleau \emph{et al.} \cite{pomerleau1989alvinn}, and it was further developed in the works \cite{lecun2005off,bojarski2016end,xu2016end}. Pomerleau \emph{et al.} \cite{pomerleau1989alvinn} attempted to use a neural network to navigate an autonomous land vehicle. With breakthrough of deep learning, DAVE-2 in \cite{bojarski2016end} learned the criterion to steer a vehicle automatically. Similarly, Xu \emph{et al.} presented a FCN-LSTM architecture in \cite{xu2016end}, which can predict egomotion of the vehicle by its previous state. All the works above lack the ability to supervise the action of the vehicle, which means we have no way to know where the vehicle is going, although the vehicle may safely drive on road.

Several control strategies using deep learning approach to control robot have been proposed in many papers.
A vision-based reinforcement learning method and evolve neural network as a controller in TORCS game have been reported in \cite{koutnik2013evolving,koutnik2013evolv}. Reinforcement learning approach in \cite{sutton1998reinforcement,mnih2015human,silver2016mastering} has been successfully used to train the artificial agent which has an capability to play several games. 
Although the combination of convolutional neural network and reinforcement learning has shown a good performance in some strategic games \cite{mnih2013playing}. This is because the decision-making in such games usually relies on a short-term of time information or the current image information. However, for complex tasks such as self-driving cars, planning and control decisions must be made with a long-term information, so as to form the optimal driving strategy for current in real traffic scene.
In \cite{chen2015deepdriving}, a direct perception approach is used to manipulate a virtual car in TORCS game. The controller in this work is a hand-crafted linear function which directly uses vehicle's position and pose. This approach may preform well in game environment, but the action generated by this function is different from human's behavior and it can not be applied in real traffic as well. Other path planning and control methods for self-driving car in \cite{paden2016survey} commonly require real time GPS information to form a real trajectory. Nevertheless, as known that a human being can drive a car only by visual information, it is a promising way to develop a model which can handle planning and control simultaneously based only on vision.

% \subsection{Our Contributions}
% Comparing with the previous works, there are four main contributions in this paper:
% \begin{itemize}
% \item It presents a novel end-to-end framework for self- driving car, which relies only on visual information, and external control instructions can be introduced into the model. 
% \item It implements a method to synchronously detect lane boundaries and obstacle vehicles rapidly, where CNN is applied.
% \item It proposes an approach which uses RNN to handle planning and controlling simultaneously, and the GPS information is not required in the whole process.
% \item It trains model with the data of human driving behaviors, and the control decisions generated by RNN is similar to the human’s, which enables the autonomous car to drive smoothly.
% \end{itemize}
%%%%%%%%%%%%方法fangfa%%%%%%%%%%%%%%%%%%%%%%%%%%%%%%

\section{Brain-inspired cognitive model with attention}
% \subsection{Cognitive Map Based Method}
In driving, human visual cortex and motor cortex play the leading roles. On the one hand, the visual cortex contributes to perceiving environment and form a cognitive map of the road scene by combining the memory of the traffic knowledge with external information, such as map navigation information. On the other hand, planning and control are determined by the motor cortex.
With the information from the cognitive map in a long memory, mechanism of attention will help people discover the most significant information in time to form planning and control strategy. 
In one word, the entire driving behavior consisting of sensing, planning and control are guided and inferred mainly by the above two cortexes in brain.

Similarly, a brain-inspired model based on the perception, memory and attention mechanisms of human can be constructed. In this paper, it is believed that the most primary perception relies on a single frame of the road scene. However, as for planning and controlling processes,  multiple frames and many historical states of vehicle are required to form long memory or short memory to  actually manipulate the self-driving car.

Fig.\ref{csttt} shows our CMA method for self-driving cars. Our ultimate goal with this network is to build a cognitive model with attention mechanism, which can handle sensing, planning and control at the same time. And differing from other deep learning or end-to-end learning methods which just map input images to uncontrollable driving decisions, our network can not only make a car run on a road, but also accept external control inputs to guide the actions of the car. To achieve this goal, the road images are first processed by multiple convolutional neural networks to simulate the function of human visual cortex and form a basic cognitive map similar to human brain's, which is a structured description of the road scene. And this description of road contains both human-defined features and latent variables learned by the network. A more explicit cognitive map can be constructed based on the contents of the basic cognitive map above and combined with prior knowledge of traffic, states of vehicle and external traffic guidance information. Therefore, the cognitive map is built by the description of the road scene, the current states of vehicle and the driving strategy of the near future. Through the recurrent neural network(RNN), the cognitive map formed in each frame is modeled to give a temporal dependency in motion control, as well as the long-term and short-term memory of past motion states, imitating human motor cortex. Finally, the real motion sequences with consideration of planning and control commands of self-driving cars can be generated.

%%%%%5
\subsection{Perception Simulating Human Visual Cortex}
\begin{figure*}[!ht]
 \centering
 \includegraphics[width=7.0in]{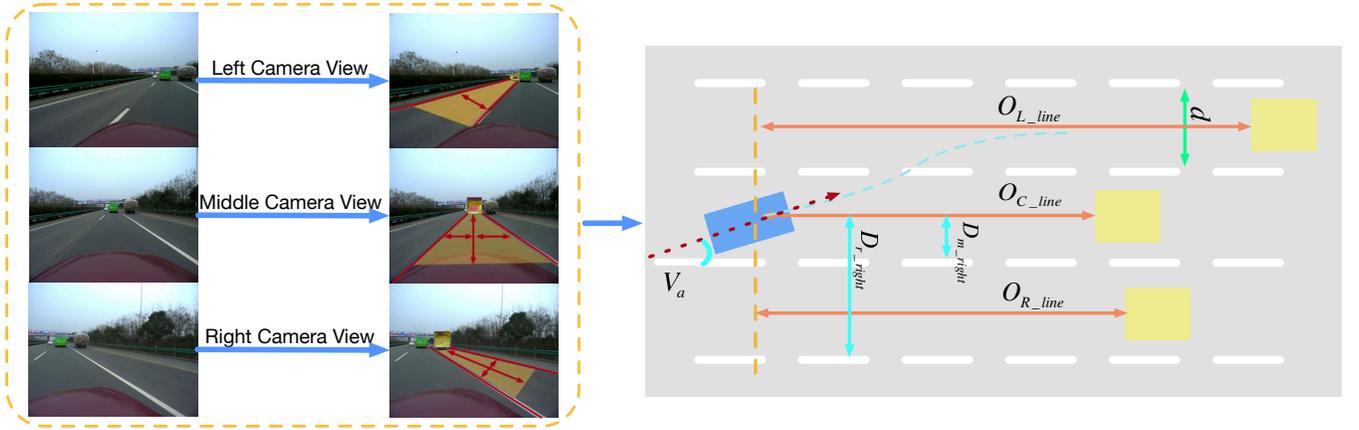}
 \caption{Illustration of constructing a basic cognitive map. The convolutional neural network will output the pixel location of lane boundary and obstacle vehicles. And based on the output, a description of local environment can be generated.}
 \label{cnnfigure1}
\end{figure*}

The purpose of the CMA framework is to solve the defect that the conventional end-to-end learning methods can not incorporate external control signals, which causes that the vehicle can only produce an action based on the input image, but does not know where it will go. By constructing a cognitive map, additional control information can be put into the end-to-end  self-driving framework. The establishment of a cognitive map primarily relies on the perception of the environment.
It is well known that the perception of environment is the focus and challenge of self-driving missions. Thus, in the CMA framework, inspired by the architecture of human visual cortex, we use a state-of-art convolutional neural network to learn and generate basic descriptions of road scene. We fixed three cameras in our self-driving car to capture the scene in  current lane and the lanes on both sides. Different from the conventional method,  the scene representations in our approach are learned by convolutional neural network but not hand-crafted. We apply several convolutional layers to process images captured by on-board cameras on vehicles. In our approach, we launch multiple convolutional networks to extract different road information from different camera views. Instead of directly using the network as a classifier, we utilize it as a regressor to directly map an input image to several key pixel points that will be used in path planning and control. With the pixel points extracted by multiple CNNs, one can calculate and construct a basic cognitive map to describe the local environment surrounding the vehicle, as shown in Fig. \ref{cnnfigure1}.

We define the input images as $\textbf{I}^t=\{I_m^t, I_l^t, I_r^t\}$, which are captured by the middle, left and right cameras, respectively. 
Based on the input images, a self-driving car needs to know the accurate geometry of the lane and position of the obstacles. So the output vector $X^t$ of the convolutional neural network for each camera is composed with five different point identities, that is
\begin{equation}
X^t = [p_{l\_t}, p_{l\_b}, p_{r\_t}, p_{r\_b}, p_o],
\end{equation}
where $p_{l\_t}$ and $p_{l\_b}$ represent the x-coordinates of the intersections of the left lane boundary's extended line with the top and the bottom edges of image plane, $p_{r\_t}$, $p_{r\_b}$ denote the x-coordinates corresponding to the points on the right lane, and the $p_o$ stands for the y-coordinate of the obstacle point in corresponding lane.

To achieve a high performance in real-time, the architecture of our convolutional neural network is very simple and shallow. Five convolutional layers are utilized to extract spatial features from each image $I^t$. The configurations of the five convolutional layers are the same as \cite{bojarski2016end}. The first three layers are stride convolutional layers with a $2\times2$ stride and a $5\times5$ kernel. While the last two convolutional layers have a $3\times3$ kernel and with no stride. Since we use the network to locate feature points rather than classify, the pooling layer which makes the representation become invariant to small translations of the input is unnecessary in our network . The convolution operation without pooling layer is expressed as
\begin{equation}\label{convlayer}
Z_{i,j,k}=\sum_{l,m,n}[V_{l,(j-1)\times{s+m},(k-1)\times{s+n}}K_{i,l,m,n}],
\end{equation}
where $Z$ and $V$ are the output feature maps with $i$ channels and input feature with $l$ channels, $s$ denotes the number of stride, $j$ and $k$ are the indexes of row and column. A rectified linear units (ReLU) is used for each hidden neurons of the convolutional layers. Following the five convolutional layers, three fully connected layers are utilized to map the representations extracted by the convolutional layers, to the output vector $X^t$.

According to the five descriptors in $X^t$, we can calculate the physical quantities as shown in Fig. \ref{cnnfigure1}.
Suppose $D_{m\_left}$ and $D_{m\_right}$ are the lateral distances to the vehicle respectively from the left and the right lane boundary in the view of the middle camera. $D_{l\_left}$, $D_{l\_right}$, $D_{r\_left}$ and $D_{r\_right}$ are respectively the two distances in the view of the left and the right camera, similar to the middle one. We define the angle between vehicle and road as $V_{a}$, and obstacle distances in each lane as $O_{C\_line}$, $O_{L\_line}$ and $O_{R\_line}$. With the obstacle distances, driving intention $D_i^t$ could be derived by Algorithm \ref{suanfa1}.
%%%推一下几个距离的公式%%%
For calculating these physical distances described above, we define that any two pixel points in a lane boundary are $(l_{xm}, l_{ym})$ and $(l_{xb}, l_{yb})$, and y-coordinate of the obstacle is $o_y$. With the optical center $(u_0, v_0)$, and the height of camera $H$, the positions of two points $(X_m,Z_m), (X_b,Z_b)$ in vehicle coordinate system can be presented as 

\begin{equation}\label{2}
(X_m,Z_m) = (\frac{(l_{xm}-v_0)\cdot{H}}{(l_{ym}-u_0)},\frac{f\cdot{H}}{l_{ym}-u_0}),
\end{equation}

\begin{equation}\label{3}
(X_b,Z_b) = (\frac{(l_{xb}-v_0)\cdot{H}}{(l_{yb}-u_0)},\frac{f\cdot{H}}{l_{yb}-u_0}).
\end{equation}
The distance $D$ from vehicle to the lane boundary can be obtained by
\begin{equation}\label{4}
D=\frac{\left | X_m Z_b-X_bZ_m \right |}
{\sqrt{(X_b-X_m)^2+(Z_b-Z_m)^2}}.
\end{equation}
Meanwhile, the angle between the vehicle and the lane boundary $V_a$ is
\begin{equation}\label{5}
V_a = \arctan \frac{X_b-X_m}{Z_b-Z_m}.
\end{equation}
Similarly, obstacle distance in each lane is presented as
\begin{equation}\label{6}
O=\frac{f\cdot{H}}{o_{y}-u_0}.
\end{equation}
By Eq. \ref{2}, \ref{3}, \ref{4}, \ref{5} and \ref{6}, we can obtain the perception results $\{X^t_m, X^t_l,X^t_r\}$ from the views of three cameras.

\begin{algorithm}[!ht]
\caption{Generating Driving Intentions with Basic Cognitive Map in Real Traffic}
\label{suanfa1}
% \label{alg::conjugateGradient}
\begin{algorithmic}
\Require \\
% $O_{C\_line}$: obstacle distance in the current lane from the middle camera\\
% $O_{L\_line}$: obstacle distance in the left lane from the left camera\\
% $O_{R\_line}$: obstacle distance in the right lane from the right camera\\
$G_{navi}$: guidance information from the navigation 
\Ensure\\
$D_i^t$: drive intention based on obstacle distances and navigation signal;
\end{algorithmic}
\begin{algorithmic}[1]
\If{$G_{navi} = \text{stay in line} \land O_{C\_line} \geq \text{safety distance}$}
\State \Return $D_i = \text{stay in line}$
\ElsIf {$G_{navi} = \text{stay in line}\land O_{C\_line} \leq \text{safety distance}$}
\If {$O_{L\_line} \geq \text{safety distance}$}
\State\Return $D_i^t = \text{change\ to\ left}$
\ElsIf{$O_{R\_line} \geq \text{safety\ distance}$}
\State\Return $D_i^t = \text{change\ to\ right}$
\Else
\State\Return $D_i^t = \text{break\ and\ stay\ in\ line} $
\EndIf
\ElsIf{$G_{navi} = \text{change\ to\ left} \land O_{L\_line} \geq \text{safety\ distance}$}
\State\Return $D_i^t = \text{change\ to\ left} $
\ElsIf{$G_{navi} = \text{change\ to\ right} \land O_{R\_line} \geq \text{safety\ distance}$}
\State\Return $D_i^t = \text{change\ to\ right} $
\Else
\State\Return $D_i^t = \text{break\ and\ stay\ in \ lane}$
\EndIf
\end{algorithmic}
\end{algorithm}
%%%%%%%%%%%%%%%%%%%%%%%%%%

%3.3%%%%%%%%%%%%%%%%%%%%%%%%%%%%%%%%%%%%%%%%%%%%%
\subsection{Planning and Control Simulating Human Motor Cortex}
The structured description of cognitive map $C^t$ with the vehicle states $V_{States}$, formed as
\begin{equation}\label{}
  C^t = 
  \begin{bmatrix}
  X_{m}^{t} & X_{l}^{t} & X_{r}^{t} & D_{i}^{t} & V_{states}
  \end{bmatrix},
\end{equation}
is extracted from each frame in three cameras. With these representations, the CMA method will model a temporal dynamical dependency of planning and control. In an ordinary self-driving framework, path planning and vehicle control are two separate tasks. Different from traditional method, the new approach contains memories of the past states and generates control commands with recurrent neural network, so the two tasks are driven simultaneously.

Long Short-Term Memory (LSTM) is a basic unit of recurrent neural network and it's well known in processing sequential data and molding temporal dependencies. 
% An LSTM unit is formed by four parts: an input gate, a forget gate, a memory cell and an output gate. 
LSTMs have many varieties, a simple one is used in our CMA framework. One cell in a LSTM unit is controlled by three gates (input gate, output gate and forget gate). Forgot gate and input gate use a sigmoid function, while output and cell state are transformed with $tanh$.
With these gates, LSTM network is able to learn long-term dependencies in sequential data and model the attention mechanism in time. We employ the outstanding characteristics of LSTM in our CMA framework, so that it can learn human's behaviors during a long-term driving process. The memory cell is used to store information in time, such as historical vehicle states in its vectors, which can be chosen with attention mechanism in the network. 
Meanwhile, the dimension of the hidden states should be chosen according to the input representation $C^t$.

Based on the driving intention $D_i$ in cognitive map $C^t$, a lateral distance $D_o$ from the current point to the objective lane is calculated by Algorithm \ref{suanfarnn}. Then a more complicated cognitive map $\textbf{C}^t$ is built as
\begin{equation}
  \textbf{C}^t = 
  \begin{bmatrix}
  X_{m}^{t}, & X_{l}^{t}, & X_{r}^{t}, & D_{i}^{t}, & V_{states}^{t}, & D_{o}^{t}
  \end{bmatrix}.
\end{equation}
Before importing the representation of cognitive map $\textbf{C}^t$ to the LSTM block, a fully connected layer is utilized to organize the various information in $\textbf{C}^t$. The representation $\textbf{R}^t$ can be presented by the fully connected layer with a weight matrix $\textbf{W}_{f}^{C}$ and a bias vector $\textbf{b}_{f}^{C}$ as
\begin{equation}
{\textbf{R}^{t}}= \textbf{W}_{f}^{c} \textbf{C}^t+\textbf{b}_{f}^{c}.
\end{equation}

The descriptor $\textbf{R}^{t}$ contains much latent information organized by the fully connected layer, which means a description of traffic scene and driving intentions. The effectiveness of this descriptor can be improved by an appropriate training process with human driving behaviors.

In the proposed CMA framework, we adopt three LSTM layers, making the network learn higher-level temporal representations. The first two LSTM blocks return their full output sequences, but the last one only returns the last step in its output sequence, so as to map the long input sequence into a single control vector to vehicles. The mapping $\Phi^R$ contains three LSTM layers with parameters $\Theta^R$ to explore temporal clues in the representation set 
\begin{equation}
\{\textbf{R}^{t}\}, t=1,2,...,n,
\end{equation} 
which contains multiple cognitive results in different time steps. The hidden state $h_3^t$ is the $t$th output of the third layer and it shows the result of a temporal model which processes the tasks of path planning and control. The hidden state  $h_3^t$ is presented as
\begin{equation}
h_3^t = \Phi^R(\Theta^R, \{h^t\}, \{\textbf{R}^{t}\}),
\end{equation}
where $\{h^t\},t=1,2,...,n$, is the hidden states set of each LSTM layer. Then, a fully connected layer defined by weights $\textbf{W}^R$ and bias $\textbf{b}^R$ will map the hidden state $h_3^t$ to a driving decision used to control self-driving cars. In a word, with the memory and predictive abilities in an LSTM block, we consider planning and controlling process as a regression problem. 

In automatic driving mode, steering command and velocity command are used to control the vehicles. In our self-driving framework, we generate these two commands separately. For the velocity command, it will be determined by traffic knowledge and traffic rules. The real velocity of the vehicle will be treat as a part of vehicle state $V_{states}^{t}$ to form cognitive map of current time step. According to a long-term cognitive map $\{\textbf{C}^t\}$, steering angle $\textbf{S}_a$ is generated by {}
\begin{equation}
\textbf{S}_a = \textbf{W}^R\cdot\Phi^R(\Theta^R, \{h^t\}, \{\textbf{W}_{f}^{c} \textbf{C}^t+\textbf{b}_{f}^{c}\}) + \textbf{b}^R,
\end{equation}
which is described above in detail.

Suppose $D_{m\_left}$ and $D_{m\_right}$ are respectively the lateral distance from the left and the right lane boundary to vehicle in the view of middle camera. And $D_{l\_left}$, $D_{l\_right}$, $D_{r\_left}$ and $D_{r\_right}$ are respectively the two distances in the left and right camera views, same as the middle one. We define the angle between vehicle and road as $V_{a}$. And we use $V_{states}$ as a representation of the states of vehicle which can be obtained through OBD port in vehicle.

With these notations, the entire workflow of CMA framework is summarized in Algorithm \ref{suanfarnn}. In a self-driving procedure, cognitive map is first constructed with the multiple convolutional neural networks. Subsequently, based on the cognitive map and vehicle states in a period of time, a final control command will be generated by recurrent neural network.

\begin{algorithm}[ht]
\caption{Planning and Control Processes by Cognitive Map and LSTM}
\label{suanfarnn}
%%%%%%%%%%%%%%%%%%%%%%%input dingyi%%%%%%%%%%%
% \begin{algorithmic}
% \Require \\
% $D_{m\_left}$: lateral distance from the left lane boundary to vehicle in the middle camera view;\\
% $D_{m\_right}$: lateral distance from the right lane boundary to vehicle in the middle camera view;\\
% $D_{l\_left}$:lateral distance from the left lane boundary to vehicle in the left camera view;\\
% $D_{l\_right}$:lateral distance from the right lane boundary to vehicle in the left camera view;\\
% $D_{r\_left}$:lateral distance from the left lane boundary to vehicle in the right camera view;\\
% $D_{r\_right}$:lateral distance from the right lane boundary to the vehicle in the right camera view;\\
% $D_i$: drive intention based on obstacle and navigation signal;\\
% $V_a$: angle between vehicle and road\\
% (eyery item above could be gotten or calculated with multiple CNNs outputs as shown in equations?????)\\
% $V_{stats}$: vehicle states including velocity, attitude and etc.\\
% $D_o$: lateral distance from the current point to the objective lane;
% \end{algorithmic}
%%%%%%%%%%%%%%%%%%%%%%%%%%%input dingyi%%%%%%%%%%%
\begin{algorithmic}[1]

\While{in self-driving mode}
\State generate cognitive map $\textbf{C}^t$ with multiple CNNs
\If{$D_{i} = stay\ in\ line$}
\State $D_o=\frac{D_{m\_left}+D_{m\_right}}{2}$
\State generate $steering\ angle$ by RNN
\State generate $pedal\ commands$ by desired speed
\ElsIf{$D_{i} = change\ to\ right$}
\State $d=D_{r\_right}-D_{r\_left}$
\While{changing lanes}
\If{$D_{m\_right} < D_{r\_right}$}
\State the car still in current lane
\State $D_o=D_{r\_right}-\frac{d}{2}$
\State generate $steering\ angle$ by RNN
\State generate $pedal\ commands$ by desired speed
\ElsIf{$D_{m\_right} == D_{r\_right}\lor \text{the car is on the boundary of lane}$}
\If{$(D_{m\_right}-D_{m\_light})\gg d$}
\State the car is on the boundary of lane
\State $D_o=D_{m\_right}-\frac{d}{2}$
\State generate $steering\ angle$ by RNN
\State generate $pedal\ commands$ by desired speed
\Else 
\State the car has changed lane
\State $D_{i} = stay\ in\ line$
\State$break$
\EndIf
\EndIf
\EndWhile
\ElsIf{$D_{i} = change\ to\ left$}
\While{changing lanes}
\State similar to change to right lane
\EndWhile
\EndIf
\EndWhile
\end{algorithmic}
\end{algorithm}

%%%%%%%%%%%%%%%%%%%%%%%%%%%%%%%%%%%%%%%%%%%%%
\section{Data Collection and Dataset} 
\begin{figure}[!ht]
% \centering
\subfloat[]{\includegraphics[width=\linewidth,trim=0 50 0 50,clip]{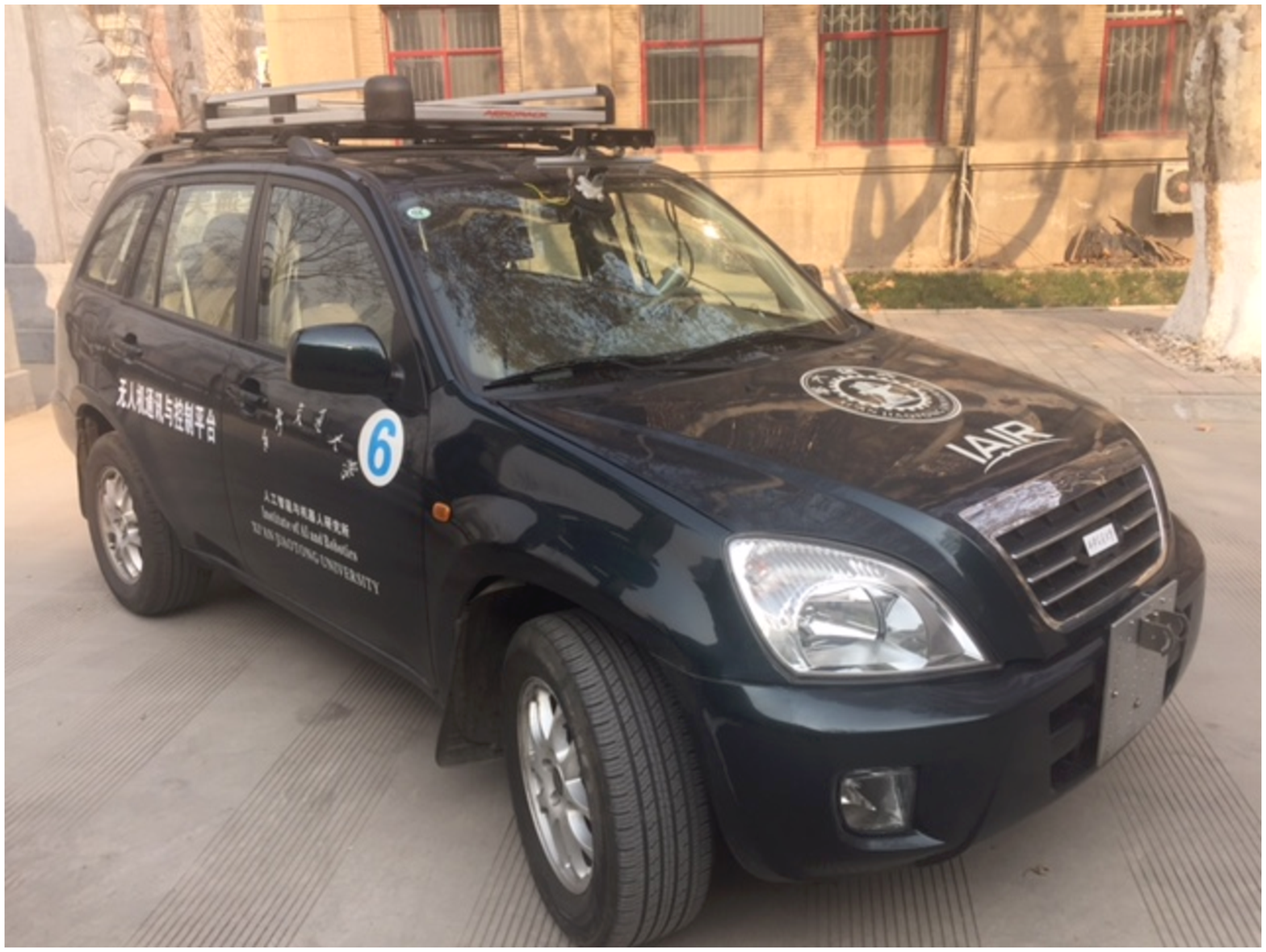}\label{sensorsetupd}}
\vfill
\begin{minipage}[]{\linewidth}
\begin{minipage}[]{0.5\linewidth}
\subfloat[]{\includegraphics[width=\linewidth,trim=0 100 0 100,clip]{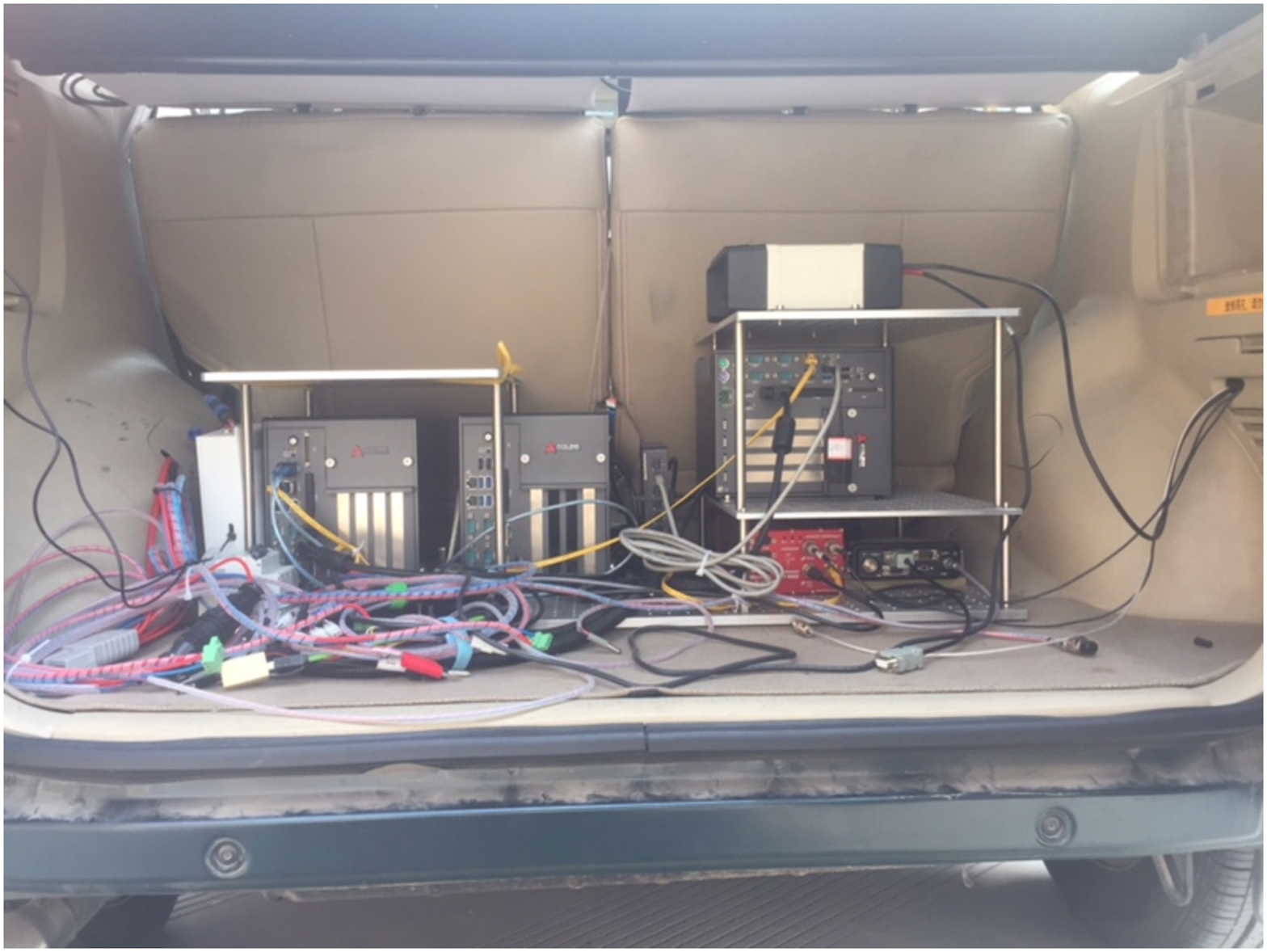}\label{sensorsetupa}}
\vfill
\vspace{-0.1in}
\subfloat[]{\includegraphics[width=\linewidth,trim=0 100 0 100,clip]{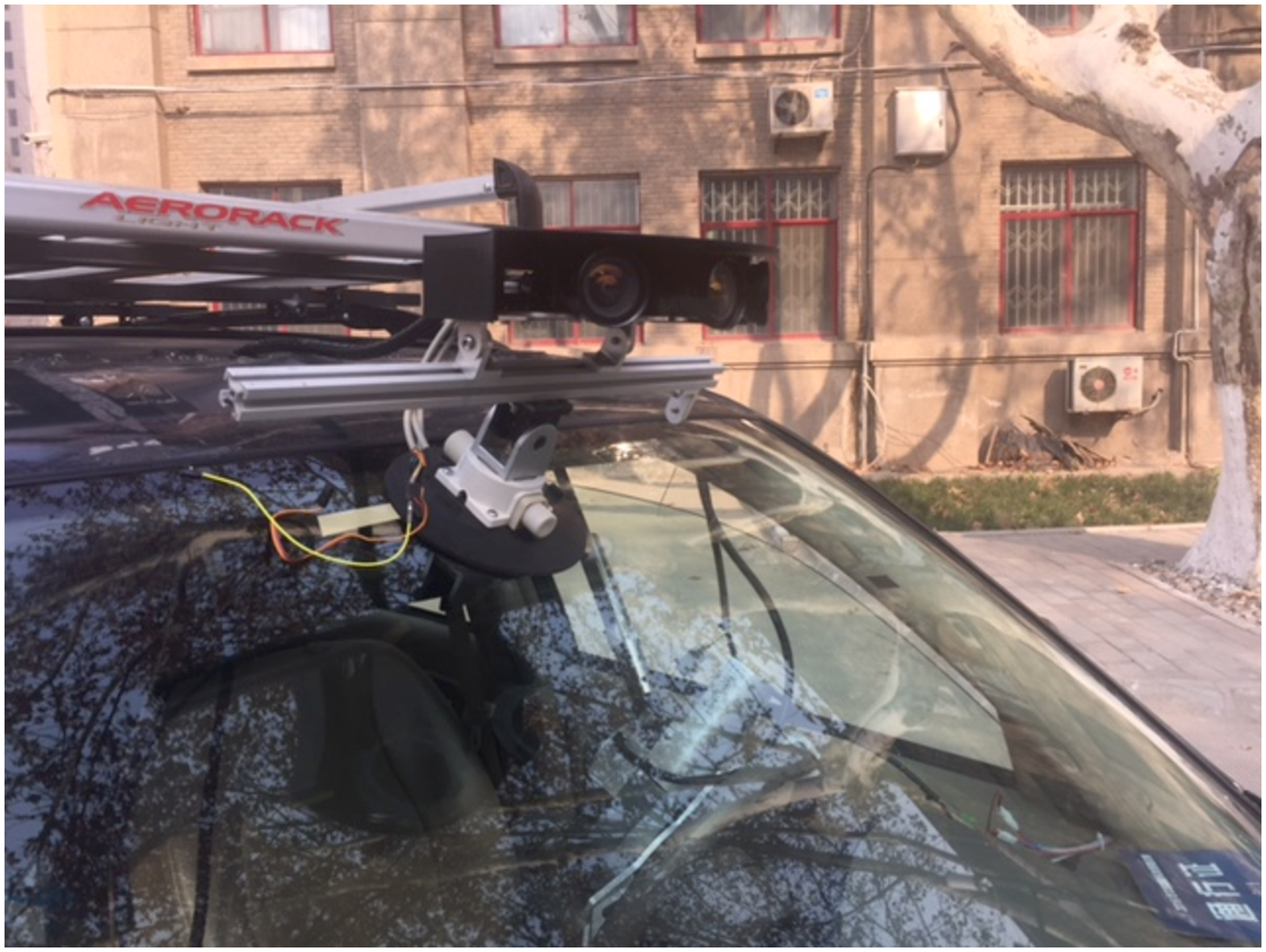}\label{sensorsetupb}}
\end{minipage}
\hfill
\begin{minipage}[]{0.455\linewidth}
\subfloat[]{\includegraphics[height=\linewidth,angle=-90,trim=50 0 50 0,clip]{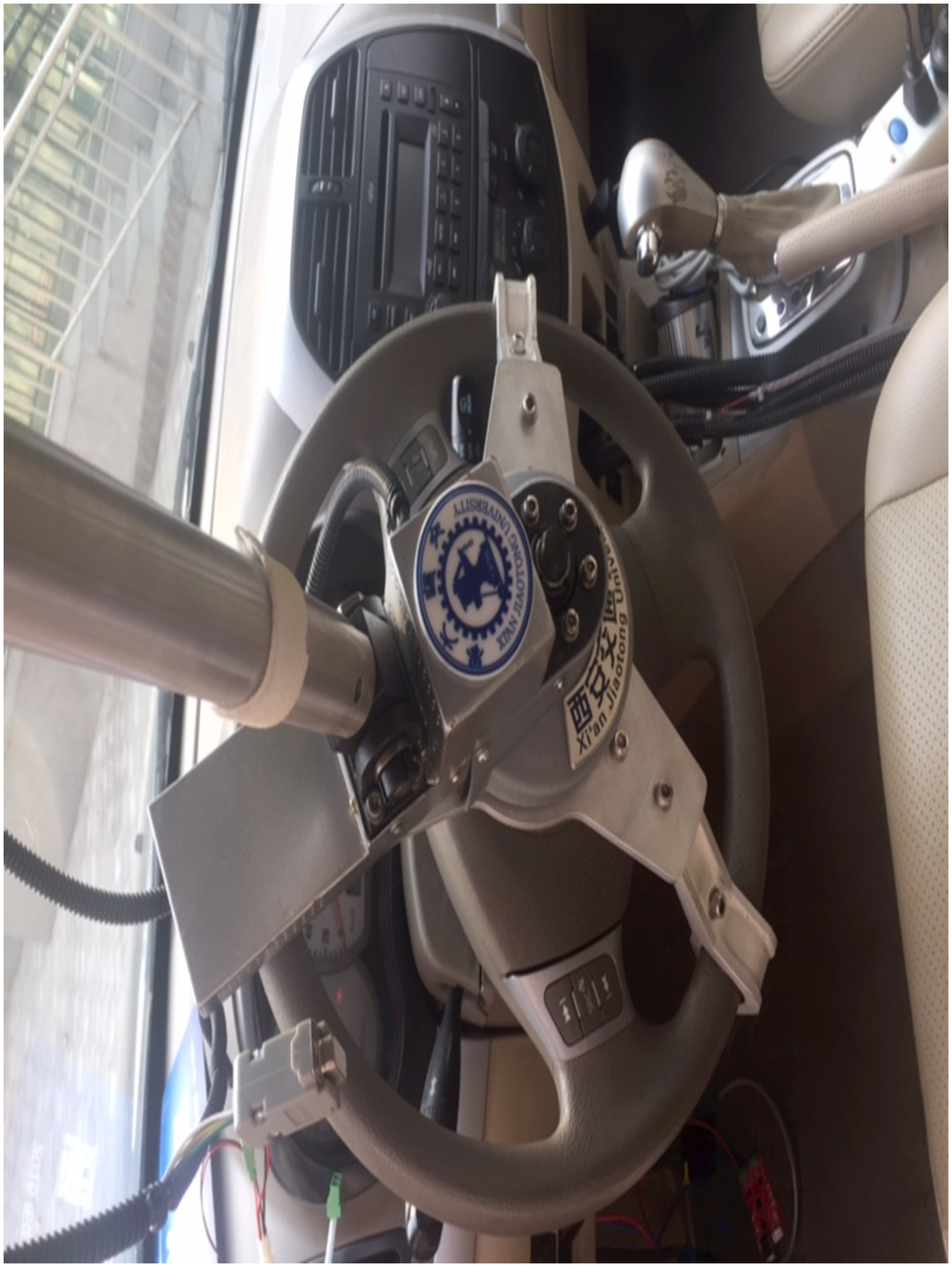}\label{sensorsetupc}}
\end{minipage}
\end{minipage}

\caption{\emph{Challenger} self-driving car. (a) Challenger is based on a \emph{CHERY TIGGO}. The \emph{TIGGO} has four-wheel drive system (4WD), and a 4-speed auto transmission. (b) Computing system and GPS/IMU inertial navigation system in the vehicle's trunk. (c) Three cameras on the top of the vehicle. Each camera monitors a corresponding lane. (d) A mechanism with a DC motor installed on the steering is used to control the steering electronically.}
\label{sensorsetup}
\end{figure}

For exploring a new framework of self-driving cars and evaluate the proposed method, we construct and publish a novel dataset: Road-vehicle Dataset (\textbf{RVD}), for training and testing our model. \footnote{The Road-vehicle Dataset is available at \emph{iair.xjtu.edu.cn/xszy/RVD.htm}}
The platform of data collection is shown in Fig. \ref{sensorsetupd}. The data is collected on a wide variety of roads in different lighting and weather conditions. 
In total, we recorded more than 10 hours of traffic scenarios using different sensors, such as color cameras, high-precision inertial navigation system and differential GPS system. Three cameras are used to record images and the data of driver's behaviors, which are reflected through the steering angle and pedals' states, are recorded by the CAN bus of OBD interface. The GPS/IMU inertial navigation system is used to recorded accurate attitude and position of our vehicle. Meanwhile, in order to generate images at different views, we use viewpoint transformation method to augment our dataset.

\subsection{Sensors and Calibration}
The sensor setup is shown in Fig. \ref{sensorsetupa}, Fig. \ref{sensorsetupb}, and Fig. \ref{sensorsetupc}:\\
\begin{itemize}
\item 3 $\times$ IDS UI-5240CP color cameras, 1.31 Megapixels, 1/1.8" e2v CMOS, global shutter
\item 3 $\times$ KOWA LM3NCM megapixel lens, 3.5mm, horizontal angle of view 89.00$^{\circ}$, vertical angle of view 73.8$^{\circ}$
\item 1 $\times$ NovAtel ProPak6 Triple-Frequency GNSS Receiver, TERRASTAR-C accuracy 4cm
\item 1 $\times$ OXTS RT2000 inertial and GNSS navigation system, 6 axis, 100Hz
\end{itemize}

For calibrating the extrinsic and intrinsic parameters of the three vehicle mounted cameras, we use trilinear method \cite{li2004calibration} to calibrate extrinsic parameters and method proposed by Zhang et.al \cite{zhang2000flexible} to calibrate intrinsic parameters. The position matrix is given as $T$, pitch angle matrix is given by $R$, and internal parameters matrix is given by $I$. And these parameters will be used later in the experiment part.
\subsection{RVD Dataset}
In the data acquisition process, drivers were asked to drive in a diverse set of weather and road conditions at different times of the day. Furthermore, to collect abundant data of driving behaviors, we required the drivers to do lane changing and turning operations in suitable cases.
For the mission of self-driving, it is a key problem to make the vehicle recover from error states. Therefore, with the method of viewpoint transformation, we combined the data from three cameras to simulate the visual data of the error-state vehicle, and generated additional road images in a variety of viewpoints.

In precise, our dataset covers:
\subsubsection{Diverse Visual data}Visual data including kinds of roads, such as the urban roadways and the highways of single-lane, double-lane and multi-lane. The data was collected in different weather conditions, such as day, night, sunny, cloudy, and foggy, by three viewpoint changeable cameras, which extends our data scale up to 146,980 images.
\subsubsection{Vehicle States Data}We recorded the data of real-time vehicle states while collecting the road video, where more than 100 kinds of internal and external vehicle information, such as speed, attitude and acceleration, was included.
\subsubsection{Driver behavior Data} We collected real-time behavior (operation to the vehicle) of the driver in each moment, including steering angle, control to the accelerator pedal and the brake pedal.
\subsubsection{Artificial Tagged Data}In our collected video, we manually tagged the 43,621 pieces of road image data, where the lane position, obstacle location, etc. were marked.

Our dataset is innovative in two aspects:
i)covers most of the visual data in scene of self-driving, all data are collected by three points of view simultaneously, and the dataset is expanded in the later stage by means of viewpoint transformation;
ii)contains abundant records of vehicle states data and human drivers' driving behaviors , which provide better exploration and training for the end-to-end frameworks of self-driving cars.

\section{Experiments}
In this section, experiments are presented to show the effectiveness of the proposed CMA model.
Comprehensive experiments are carried out to evaluate the cognitive map (free space, lane boundary, obstacle, etc.) formed by CNNs based on the data of real traffic scene videos.
We also evaluate the path planning and vehicle control performance of the RNN that integrates the cognitive map with the long-short term memory.
In addition, a simulation environment is set up,in which some experiments hard to operate in reality can be carried out.
% In order to assess the method comprehensively, the evaluation is carried out in two ways: one is to evaluate the CNN's capability of forming a cognitive map based on the data of real road scene video, where the performance on visual tasks such as free space detection, lane boundary detection, or obstacle detection is focused; the other is to test RNN's capability of integrating the cognitive map with the long-short term memory to finish path planning and vehicle control.
% In order to assess the method comprehensively, the evaluation is carried out in two ways: one is evaluating the CNN's capability of forming the cognitive map based on the data of real road scene video, where the performance on visual tasks such as free space detection, lane boundary detection, or obstacle detection is focused; the other is testing RNN's capability of integrating the cognitive map with the long-short term memory to finish path planning and vehicle control.
% And also, we set up a simulation environment in this part, which is so real that some experiments hard to operate in reality can be carried out in this simulated one.
All experiments are based on the \emph{challenger} self-driving car, as shown in Fig. \ref{sensorsetup}, a tuning vehicle on \emph{CHERY TIGGO}.
% In this section, experiments are presented to show the performance of the proposed CMA framework.
% In order to assess the method comprehensively, the evaluation mainly consists of two parts. 
% In the first part, our network is evaluated by its ability to construct the cognitive map based on the data of real road scene video, where the performance on visual tasks such as safe-driving area detection, lane boundary detection, or obstacle detection is focused.
% In the second part, we train the network by applying data collected from human driving processes and make the CMA learn from it. Then the network is evaluated by its capability to integrate the cognitive map with the long-short-term memory to do path planning and vehicle control. And also, we set up a simulation environment in this part, which is so real that some experiments hard to operate in reality can be carried out in this simulated one.
% All of our experiments are based on the \emph{challenger} self-driving car, as shown in Fig. \ref{sensorsetup}, which is a tuning vehicle on \emph{CHERY TIGGO}.

\subsection{Constructing Cognitive Maps with Multiple Convolutional Neural Networks}
\subsubsection{Visual Data and Visual Data Augmentation}
Training a convolutional neural network to generate the descriptions of road scenarios needs a lot of image data. Although we've already got plenty of images taken by the vehicle-mounted camera, it's still hard to cover all possible situations a self-driving vehicle may encounter. Only recording images from the driver's point of view is inadequate, so our method should have an ability to adjust the vehicle to recover from an error state. For example, the samples taken from a human-driven vehicle can not cover the situation where its yaw angle is very large, since a human driver does not allow the car to deviate too much from the middle of the road. 
Thus, the collected data may be unbalanced, and it is hard to train a network that can deal with various road situations.  
% Thus, the data captured may be unbalanced for the training process, which means the output of network may not be robust in various road situations. 
In order to augment our training data and simulate a variety of attitudes of vehicle driving on the lane, we propose a simple viewpoint transformation method to generate images with different {rotations} from the direction of the lane. We record images from three different viewpoints by three cameras mounted on the vehicle, and then simulate other viewpoints by transforming the images captured by the nearest camera. Viewpoint transformation requires the precise depth of each pixel which we cannot acquire. However in our research, we only care about the lanes on the road. We assume that every point on the road is on a horizontal plane. So one can accurately calculate the depth of each pixel on the road area by the height of camera.

\begin{figure}[!ht]
 \centering
 \includegraphics[width=3.5in,trim=0 0 0 0,clip]{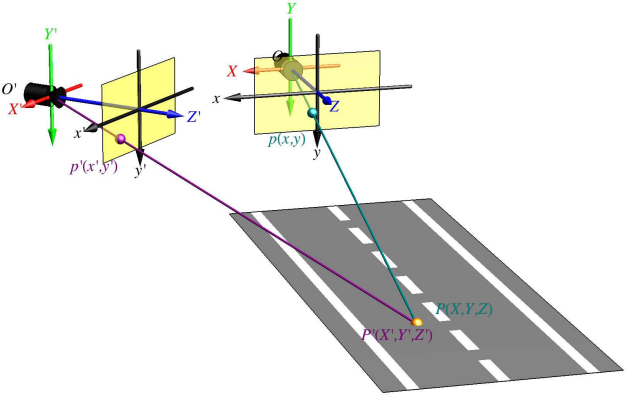}
 \caption{Principle of viewpoint transformation.}
 \label{viewpoint}
\end{figure}

\begin{figure*}
 \centering
 \includegraphics[width=7.0in]{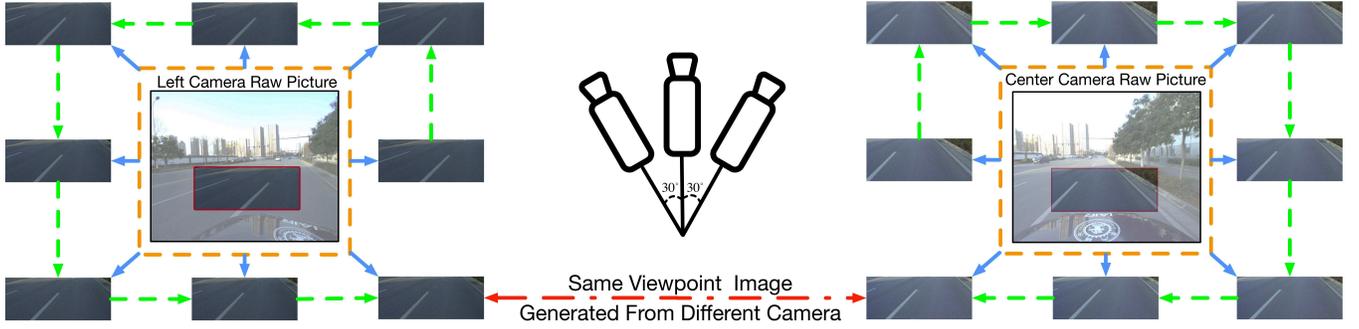}
 \caption{Illustration of data augmentation process. Images in different viewpoints are generated by the method of viewpoint transformation. The origins are the actual images taken by the three cameras.}
 \label{viewpointfig}
\end{figure*}
Suppose a point $p(x,y)$ in the image coordinate system is known, the pitch angle of the camera is approximate to $0$. With the basic camera model illustrated in Fig. \ref{viewpoint}, one can obtain the position of the point $P(X,Y,Z)$ in the camera coordinate system as
\begin{equation}
Z = Y \times \frac{f}{y},
\end{equation}
\begin{equation}
X = Z \times \frac{x}{f},
\end{equation}
where $Y$ equals the height $h$ of the camera, and the focal length of the camera is $f$.

A point $P'(X',Y',Z')$ in the simulated camera coordinate system can then be derived as
\begin{equation}
\left[                 
\begin{array}{c}   
X'\\Y'\\Z'
\end{array}
\right]
=R \times
\left[                 
\begin{array}{c}   
X\\Y\\Z
\end{array}
\right]  
+  
T,                  
\end{equation}
where $R$ is the rotation matrix and $T$ is the translation vector. Therefore, as shown in Fig. \ref{viewpointfig}, the augmented samples are generated.

\subsubsection{Effects of Multiple CNNs in Constructing Cognitive Map}
Within the proposed CMA framework, the CNN regressor takes responsibility for the construction of a basic cognitive map. In a vision based self-driving car, how precisely the visual module can perceive the surrounding environment is of particular importance, and the perception results directly affect the planning and control. In most self-driving scenarios, detections of the free space, the current and adjacent lanes, as well as the obstacle vehicles are primary indicators.  Therefore, we mainly evaluate our model on detecting these items.

\begin{table}[ht]
\renewcommand{\arraystretch}{1.2}
\caption{Parameters of the five convolutional layers}
\label{tablecnn}
\centering
\begin{tabular}{c||rl}
\hline
\makecell[c]{\bfseries Layers} &
\makecell[c]{\bfseries Operations} & 
\makecell[c]{\bfseries Attributions} \\
% \hhline{=#==}
\hline\hline
\multirow{3}{*}{1st}
& Convolution & Size: $\left[5 \times 5 \times 3 \times 24\right]$ \\
& Activation & ReLU \\
& Max pooling & (Not Used)\\
\hline
\multirow{3}{*}{2nd}
& Convolution & Size: $\left[5 \times 5 \times 24 \times 36\right]$ \\
& Activation & ReLU \\
& Max pooling & (Not Used)\\
\hline
\multirow{3}{*}{3rd}
& Convolution & Size: $\left[5 \times 5 \times 36 \times 48\right]$ \\
& Activation & ReLU \\
& Max pooling & (Not Used)\\
\hline
\multirow{3}{*}{4th}
& Convolution & Size: $\left[5 \times 5 \times 48 \times 64\right]$ \\
& Activation & ReLU \\
& Max pooling & (Not Used)\\
\hline
\multirow{3}{*}{5th}
& Convolution & Size: $\left[5 \times 5 \times 64 \times 128\right]$ \\
& Activation & ReLU \\
& Max pooling & (Not Used)\\
\hline
\end{tabular}
\end{table}

Our CNN regressor is built on TensorFlow \cite{abadi2016tensorflow}. As given in Table \ref{tablecnn}, there are 5 convolutional layers in our model. An image with resolution $320\times240$ is processed by those convolutional layers to form a tensor with dimension $25\times33\times128$. And 4 fully connected layers with output size $500, 100, 20, 5$ are used to map $128$ features to a 5-dimensional vector, which represents the lane boundary and obstacle location in the input image. 

\textbf{The performance of free-space estimation} is evaluated by a segmentation-based approach. In our approach, we can estimate the free-space in the ego-lane or the adjacent lanes. Free-space in a lane is determined by two lane boundaries and obstacle position in the lane. We evaluate the consistency between the model output and accurately labeled ground truth. Despite that in \cite{fritsch2013new}, the authors recommend to estimate the free space in bird's eye view regardless of the type of traffic scenarios, here in our model, we alternatively choose to evaluate by perspective image pixels. Since what we concern indeed is the free-space in a lane, the adopted perspective is more convenient for us. 
We choose the criteria of \emph{precision}, \emph{recall} and \emph{F-measure} to evaluate the performance of our model, which are defined by:
\begin{align}
Precision &= \frac{N_{TP}}{N_{TP} + N_{FP}}\ , \\
Recall &= \frac{N_{TP}}{N_{TP} + N_{FN}}\ , \\
F_1 &= \frac{2}{Precision^{-1} + Recall^{-1}} \notag\\
&= \frac{2 N_{TP}}{2 N_{TP} + N_{FP} + N_{FN}}\ ,
\end{align}
where $N_{TP}$ is the number of free-space pixels correctly labeled as ground truth, $N_{FP}$ is the number of free-space pixels in model output but not in the ground truth labeling, and $N_{FN}$ is the number of free-space pixels in ground truth but not in model output. In our study, three lanes (ego-lane, two adjacent lanes) are captured by three different cameras. We only present the evaluation results on the images of the middle camera and the left camera, since the images of the right camera are similar to those of the left one.

\begin{figure}[ht]
\begin{minipage}[]{\linewidth}
% \includegraphics[width=\linewidth]{zone1} 
% \vfill
% \vspace{0.05in}
% \includegraphics[width=\linewidth]{zone2} 
% \vfill
% \vspace{0.05in}
% \includegraphics[width=\linewidth]{zone3} 
% \vfill
% \vspace{0.05in}
% \includegraphics[width=\linewidth]{zone4} 
\includegraphics[width=\linewidth]{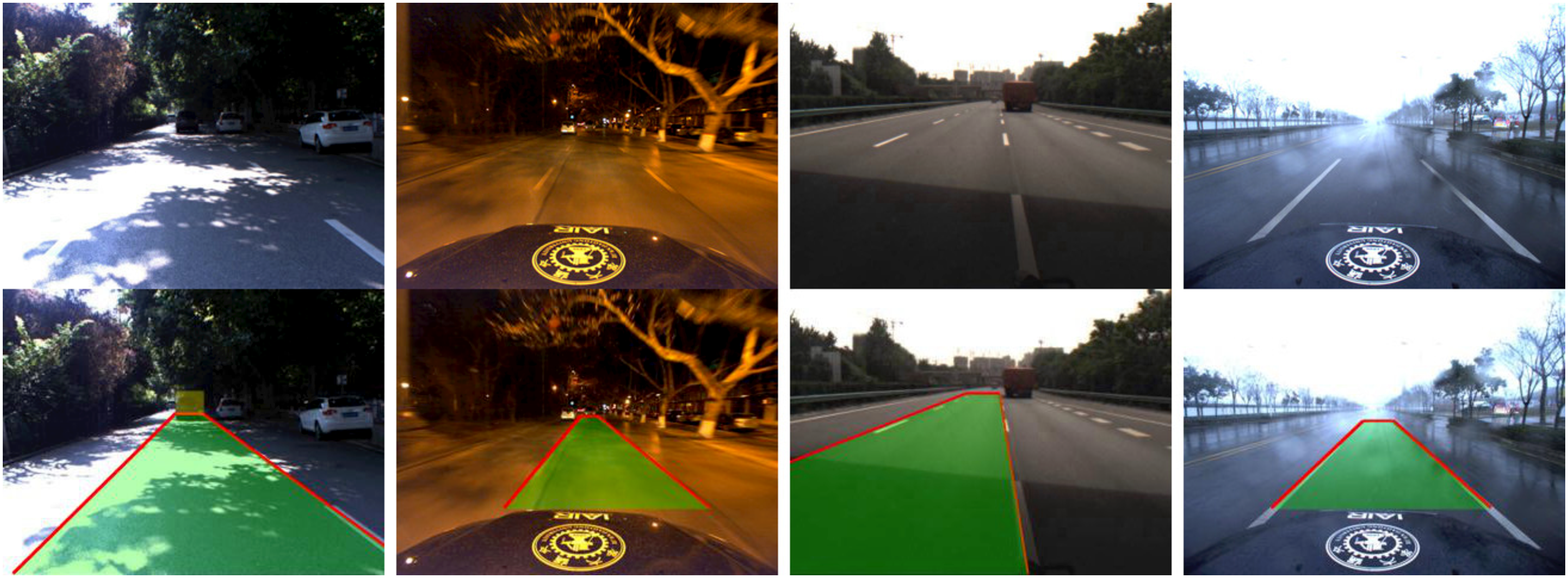} 
\vfill
\vspace{0.05in}
\includegraphics[width=\linewidth]{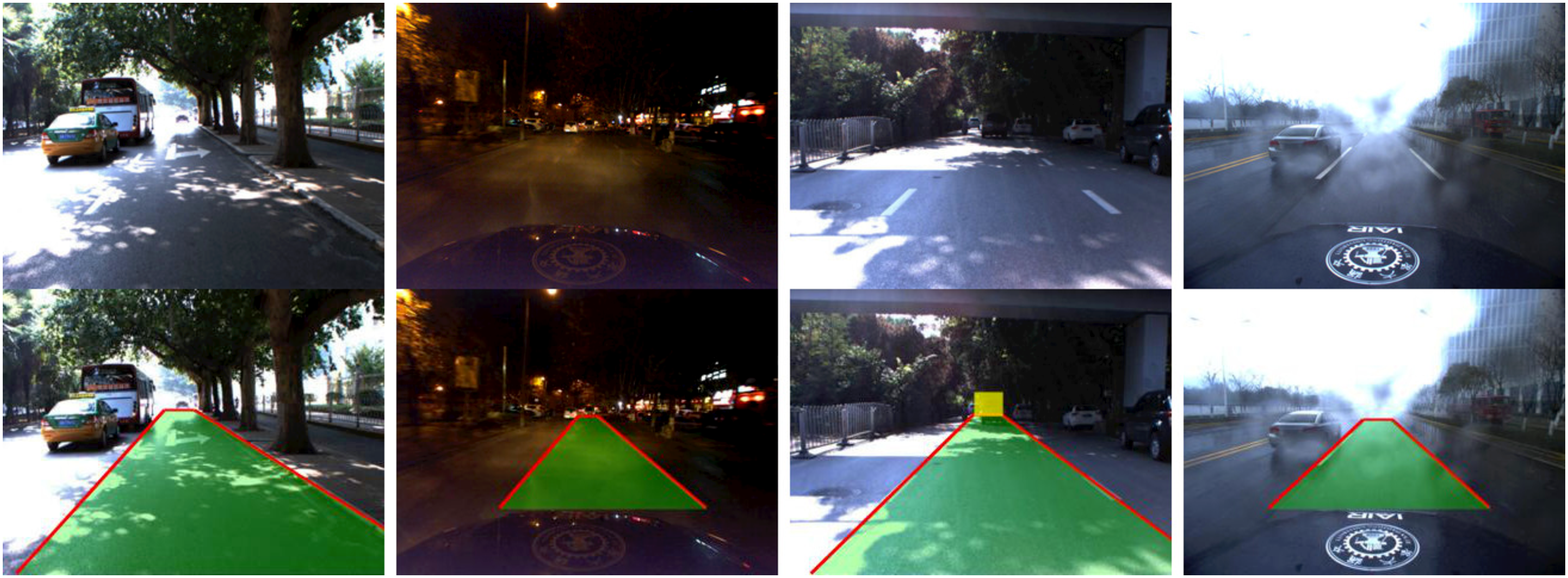} 
\vfill
\vspace{0.05in}
\includegraphics[width=\linewidth]{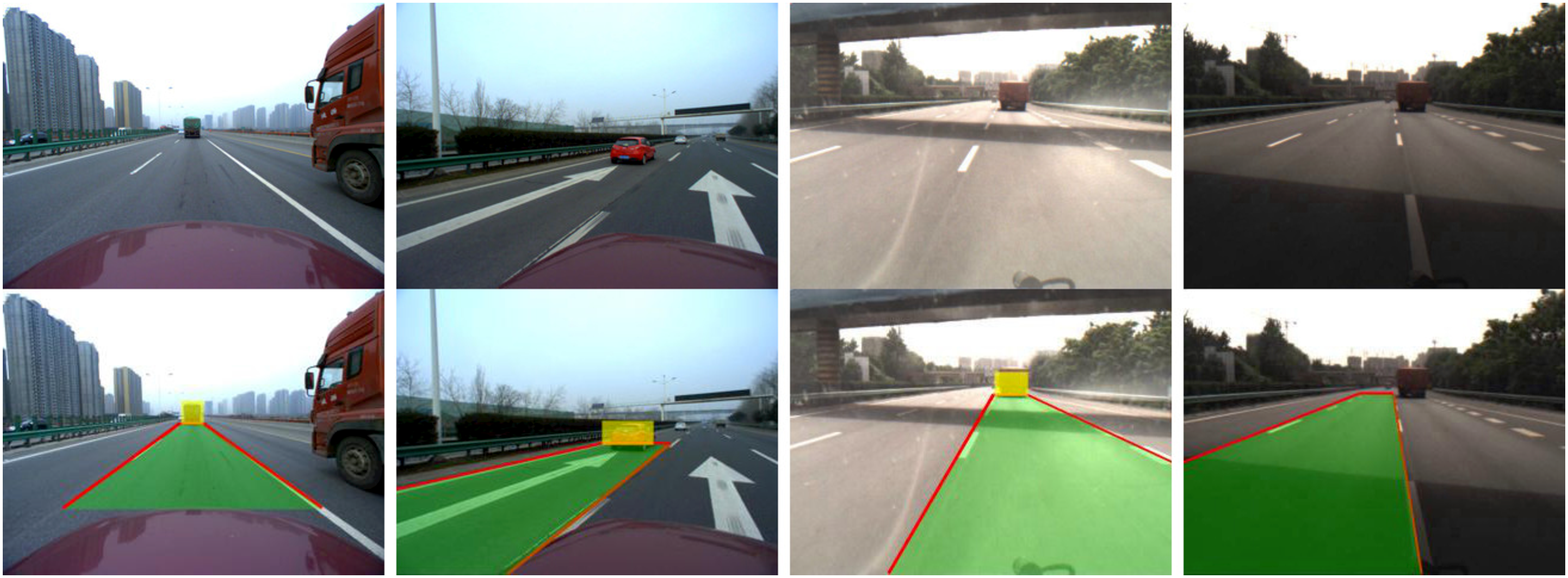} 
\vfill
\vspace{0.05in}
\includegraphics[width=\linewidth]{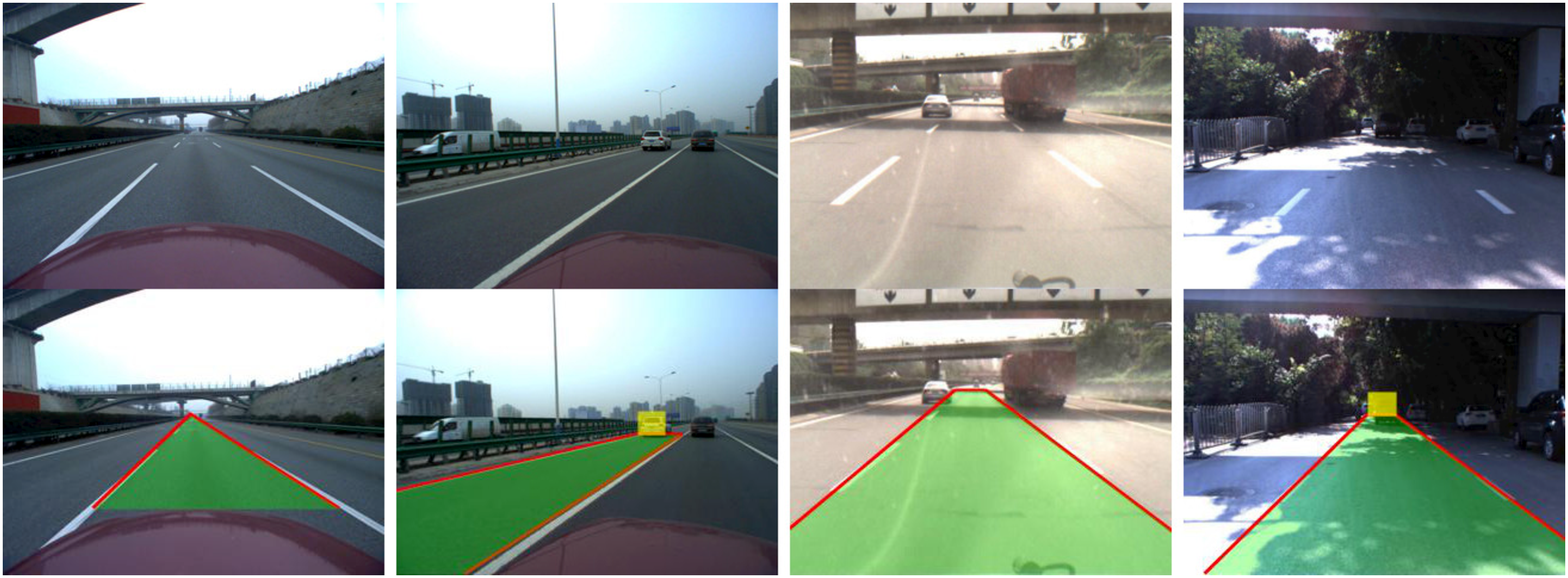} 
% (e) & \includegraphics[width=\linewidth]{lt457} \\
\end{minipage}
\caption{Free-space detection results. The proposed model is evaluated on our RVD dataset and Nan's \cite{nan2016efficient}, where various traffic scenes such as night, rainy day and complex illumination are included.}
\label{figurefreespace}
\end{figure}

For comparison purpose, the free space detection performance is evaluated on the RVD and Nan's dataset \cite{nan2016efficient}, respectively.
% To evaluate the performance of free space detection comparatively, the evaluations are respectively carried on RVD dataset and Nan's proposed dataset \cite{nan2016efficient}. %%%对于检测可行驶区域的性能评估，我们在我们提出的RVD数据集和Nan提出的数据集上进行%%%
Table \ref{tablefreespace} presents the quantitative analysis of our model in different scenarios. Some results of free-space detection on testing set are shown in Fig. \ref{figurefreespace}.

\begin{table}[ht]
\renewcommand{\arraystretch}{1.3}
\caption{The performance of free-space detection in our RVD dataset and nan's \cite{nan2016efficient}} 
\label{tablefreespace}
\centering
\begin{tabular}{c||cccc}
\hline
\bfseries Traffic Scene & 
$\overline{Precision}[\%]$ &
$\overline{Recall}[\%]$ &
$\overline{F_1}[\%]$ \\
\hline\hline
Urban & 98.16 & 97.51 & 97.82 \\
Moderate Urban & 98.33 & 97.68 & 97.98 \\
Complex Illumination & 97.88 & 99.45 & 98.65 \\
HighWay & 98.97 & 98.94 & 98.95 \\
HighWay (Left Lane) & 99.48 & 92.02 & 95.59 \\
Cloudy & 98.33 & 97.68 & 97.98 \\
Rainy \& Snowy Day& 96.83 & 97.99 & 97.37 \\
Night & 97.68 & 97.70 & 97.66\\
Heavy Urban \cite{nan2016efficient} & 98.32 & 96.91 & 97.58 \\
Highway \cite{nan2016efficient} & 99.24 & 98.43 & 98.83 \\
\hline
\end{tabular}
\end{table}

% \begin{figure}[ht]
% \begin{tabular}{m{0px}m{0.93\linewidth}}
% \setlength{\tabcolsep}{0px}
% (a) & \includegraphics[width=\linewidth]{demo1} \\
% (b) & \includegraphics[width=\linewidth]{demo1} \\
% (c) & \includegraphics[width=\linewidth]{demo1} \\
% (d) & \includegraphics[width=\linewidth]{demo1} \\
% % (e) & \includegraphics[width=\linewidth]{lt457} \\
% \end{tabular}
% \caption{Free-space detection result.}
% \label{figurefreespace}
% \end{figure}

\textbf{The results of lane boundary detection} is evaluated with the criteria presented in \cite{nan2016efficient}. If the horizontal distance between detected boundary and ground truth labeling is smaller than a predefined thresholds, the detected boundary will be regard as true positive.
%%%%%叙述我们的结果有多么的好，在很多challenging的情况下表现的很好，包括雨天，雪天，雾霾天,模型对不同车道进行检测，我们的模型我们对比了我们的方法与nan的方法，结果显示，在不同的场景下，我们模型的准确度与nan相当，但是由于通过了GPU加速，所以帧率高于nan的结果。特别是在雨雪天气，我们的模型显示出了更强的鲁棒性%%%%%%
%我们将我们提出的方法与state-of-art的方法进行了比较。在nan提出的数据集与我们的RVD数据集中，将两种方法进行了对比评估。 结果显示，我们的方法在一般的道路场景中，与state-of-art的方法展现出了近乎一致的precision，有着相似的表现。然而在challenging 的scenarios中，如雨+学+雾霾的情况下，我们的模型表现出了更好的性能，说明了我们的模型有比较强的鲁棒性robust。同时，我们的模型由于采用了convolutional neural network，所以可以在GPU上进行并行运算，所以帧率相较于nan’s方法更高%
We compared our approach with the state-of-art method. The two methods were evaluated respectively in our RVD dataset and Nan's dataset \cite{nan2016efficient}. The experimental results show that our approach exhibits a nearly consistent precision with the state-of-art approach in some typical traffic scenes. However, in challenging scenarios, such as in rainy, snowy and hazy day, our model shows better performance, indicating that it has a relatively strong robustness. Additionally, with the utilization of convolutional neural network, our model can be processed in parallel on GPU, which leads to a higher frame rate comparing with the state-of-art method.
\begin{table}
\renewcommand{\arraystretch}{1.3}
\caption{Comparison with other lane boundary detection methods in different traffic scenes}
\label{tablelaneboundary}
\centering
\begin{tabular}{cc||ccc}
\hline
\bfseries Traffic Scenes &
\bfseries Methods &
$Precision[\%]$ &
Frame Rate \\
\hline\hline
\multirow{2}{*}{Highway}
& Ours & 99.9 & 93 \\
& Nan's & 99.9 & 28 \\
\hline
\multirow{2}{*}{Moderate Urban}
& Ours & 96.2 & 90 \\
& Nan's & 97.7 & 21 \\
\hline
\multirow{2}{*}{Heavy Urban}
& Ours & 96.1 & 90 \\
& Nan's & 95.4 & 16 \\
\hline
\multirow{2}{*}{Illumination}
& Ours & 95.8 & 89 \\
& Nan's & 100.0 & 23 \\
\hline
\multirow{2}{*}{Night}
& Ours & 90.7 & 94 \\
& Nan's & 99.4 & 35 \\
\hline

% \multirow{2}{*}{Heavy traffic}
% & Ours & OOO & 100 \\
% & N's & OOO & 100 \\
% \hline
% \multirow{2}{*}{Clear Road}
% & Ours & OOO & 100 \\
% & N's & OOO & 100 \\
% \hline
% \multirow{2}{*}{Highway(XZ)}
% & Ours & OOO & 100 \\
% & N's & OOO & 100 \\
% \hline
\multirow{2}{*}{Rainy \& Snowy Day }
& Ours & 87.1 & 89 \\
& Nan's & 47.3 & 21 \\
\hline
\end{tabular}
\end{table}

\begin{figure}[ht]
% \begin{tabular}{m{0px}m{0.93\linewidth}}
\begin{minipage}[]{\linewidth}
% \setlength{\tabcolsep}{0px}
% (a) & \includegraphics[width=\linewidth]{lane1} \\
% (b) & \includegraphics[width=\linewidth]{lane2} \\
% (c) & \includegraphics[width=\linewidth]{lane3} \\
% (d) & \includegraphics[width=\linewidth]{lane4} \\
% \includegraphics[width=\linewidth]{lane1} 
% \vfill
% \vspace{0.05in}
% \includegraphics[width=\linewidth]{lane2}
% \vfill
% \vspace{0.05in} 
% \includegraphics[width=\linewidth]{lane3} 
% \vfill
% \vspace{0.05in}
% \includegraphics[width=\linewidth]{lane4} 
\includegraphics[width=\linewidth]{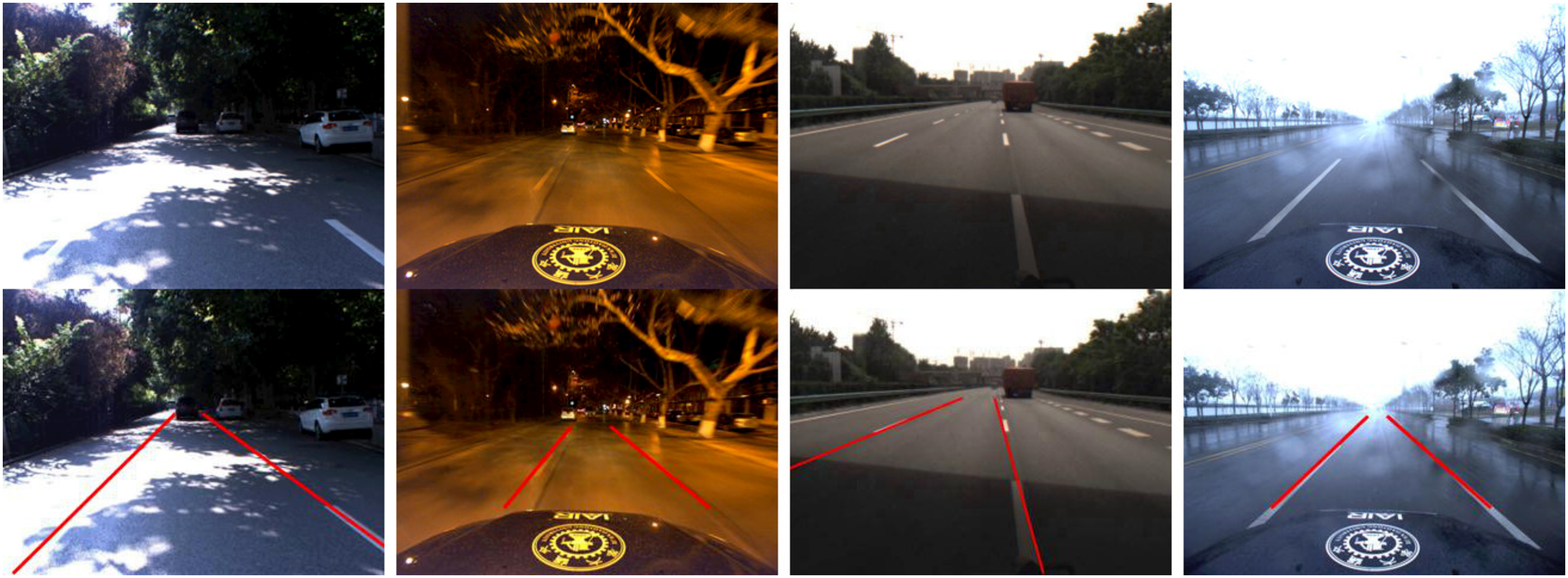} 
\vfill
\vspace{0.05in}
\includegraphics[width=\linewidth]{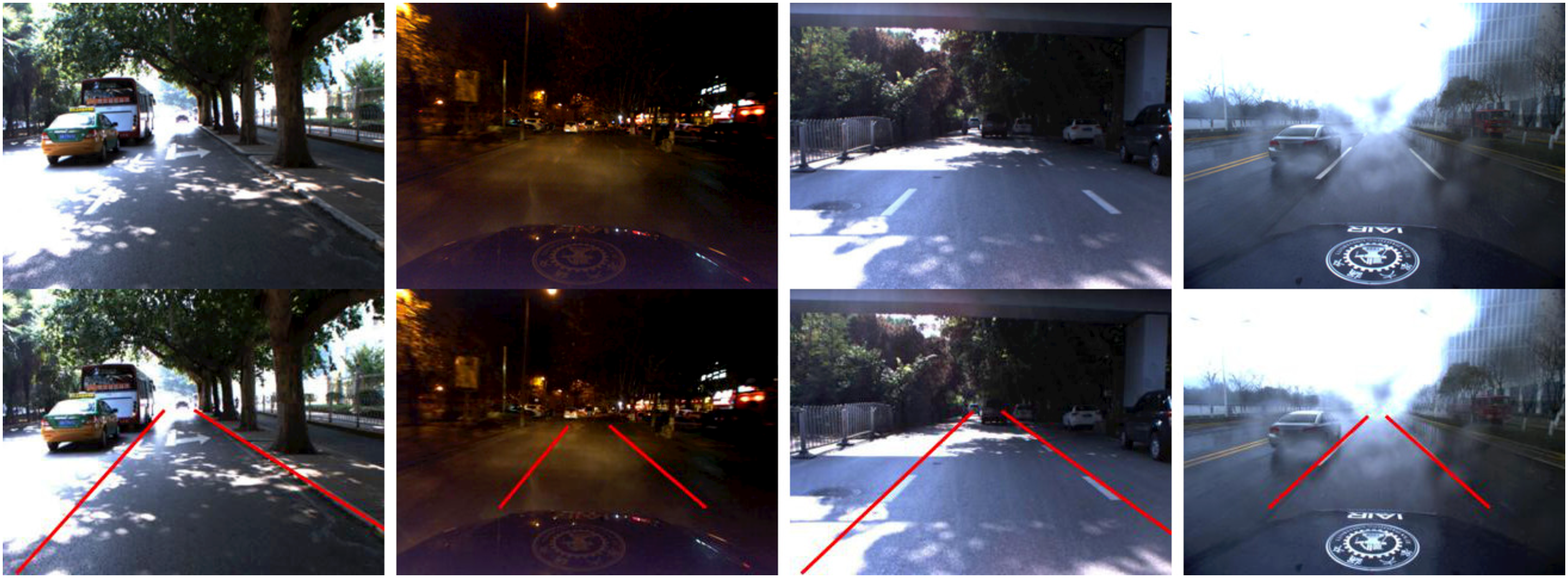}
\vfill
\vspace{0.05in} 
\includegraphics[width=\linewidth]{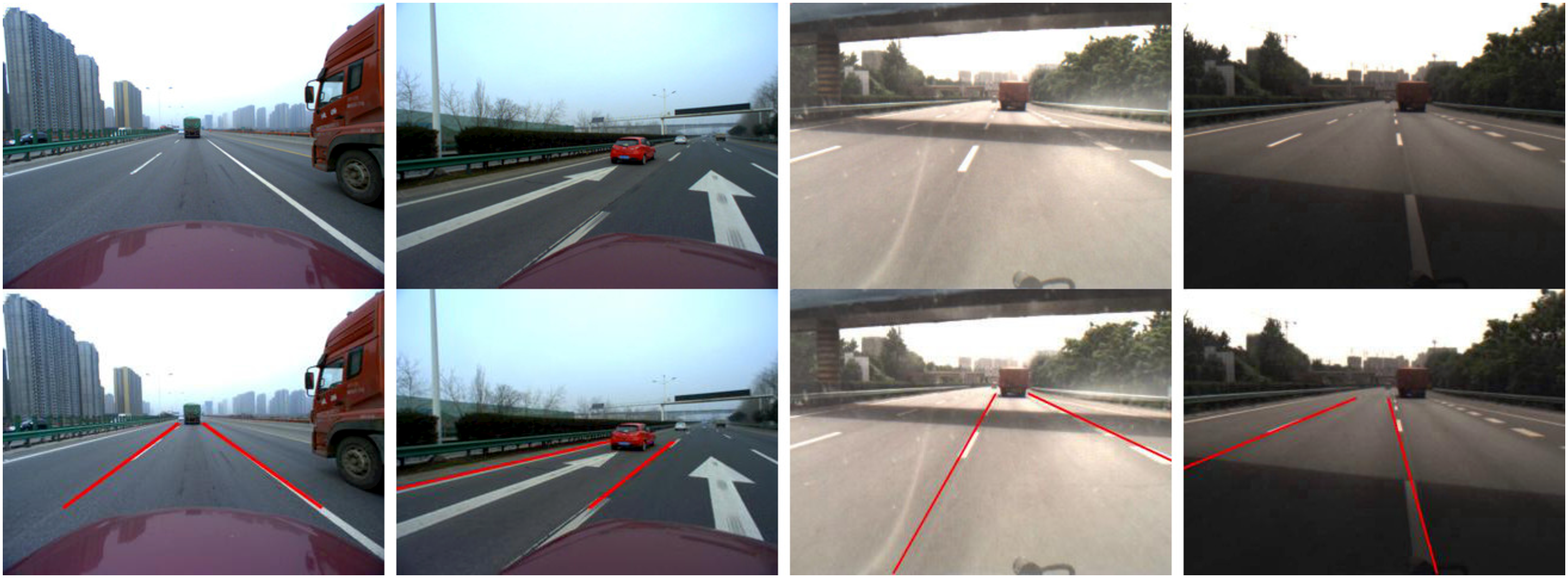} 
\vfill
\vspace{0.05in}
\includegraphics[width=\linewidth]{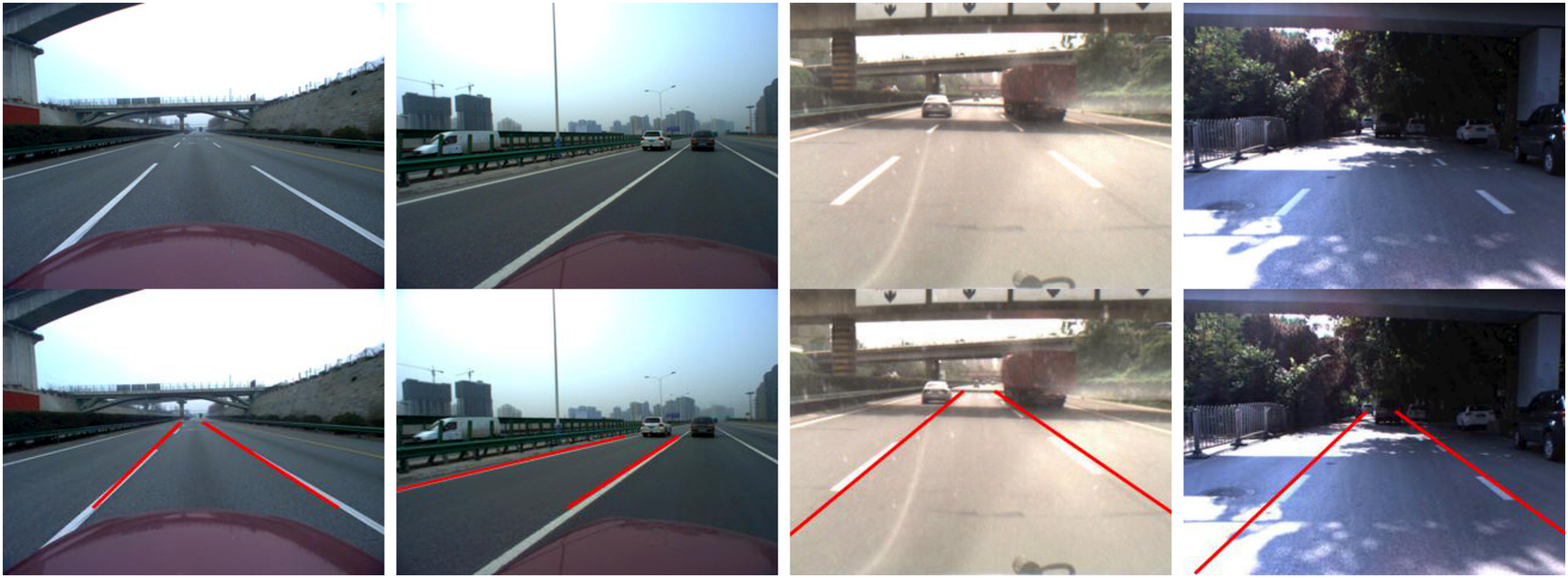} 
\end{minipage}
\caption{Lane boundary detection results on different datasets.}
\label{figurelaneboundary}
\end{figure}

The quantitative results of our model in lane boundary detection are presented in Table \ref{tablelaneboundary}. 
The lane boundary detection results of our model in different scenes and datasets are demonstrated in Fig. \ref{figurelaneboundary}.
% And as shown in Fig. \ref{figurelaneboundary}, some example figures of our model outputs in different scenes and different datasets are presented.

\begin{figure}[!htb]
 \centering
 \includegraphics[width=3.5in]{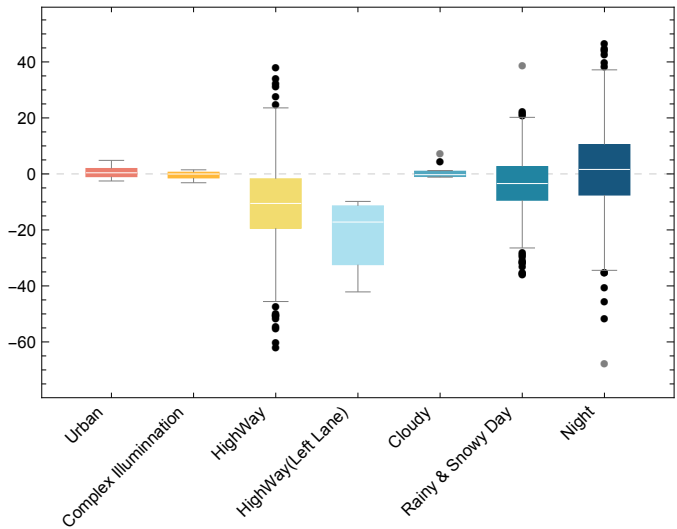}
 \caption{Distributions of the distance errors.}
 \label{figure_obstacle_distance}
\end{figure}
\textbf{The estimation of obstacle position} with our model is evaluated in perspective image pixels, since the real distance between obstacle vehicle and our car is related to extrinsic parameters of camera which may differ in different self-driving cars. At first we calculate the accuracy of the obstacle detection. 
% As shown in Table. ,the accuracy of detecting the obstacle is present, and Fig. \ref{figure_obstacle_distance} is the error distribution of distance precision in different scenes.
The distributions of the distance errors in different scenes are shown in Fig. \ref{figure_obstacle_distance}.

%%%%%%%%%%%%%%%%%%%%%%%%%%%%%%%%%%%%%%%%%%%%%%%%%%%%%%%
\subsection{Generating Control Command Sequences with Recurrent Neural Network}
\subsubsection{Experiment Setup}
\begin{figure}[htb]
 \centering
 \includegraphics[width=3.5in]{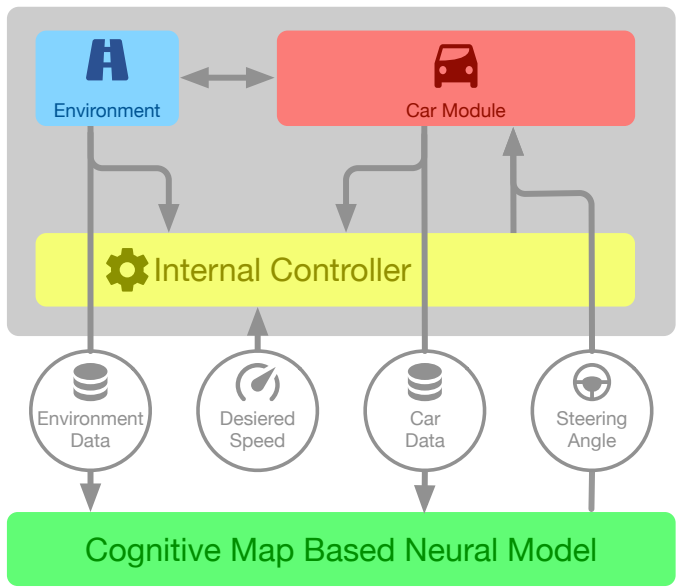}
 \caption{The block diagram of our simulator, which is used to evaluate the performance of our CMA model in path planning and control. The simulation system mainly consists of three parts, which are internal controller, vehicle model and road circumstance.}
 \label{simulator}
\end{figure}

As shown in Fig. \ref{simulator}, the simulation environment we constructed mainly contains three parts, which are road circumstance, vehicle model and internal controller (driver model included). There are two main proposes for our simulator, one is to evaluate the performance of our method, the other is to augment the data of driver behaviors. In our simulator, the car model interacts with the road circumstance, and their status are sent to the internal controller as inputs to simulate human driver's behaviors, so as to generate control sequences to adjust the attitude of the car. These three modules constitute a complete closed-loop simulation cycle.

\subsubsection{Effects of Control Sequence Generation} 
Training a recurrent neural network to learn human's driving behavior needs a large-scale driving behavior data. However, lots of unexpected factors may take effect on the data we get from the real world driving. For example, a human's driving behavior is partially based on his subjective decision, which may be absolutely different even in a same circumstance. Existences of such things in our training data may lower the confidence of our network's output. Therefore, to evaluate the planning and control part, we use the data not only from our dataset, which contains both road data we marked and actions the driver actually made, but also from the simulation environment, shown in Fig. \ref{simulator}. By setting the parameters of both the vehicle and the road module properly, the desired driving data can be reliably generated from the simulation environment, such as the driving trajectory and the commands of steering the wheel of the car. We set multiple parameters for the road module, and run the simulator repeatedly, so as to obtain information of vehicle states (vehicle attitude, speed etc.) and deviation between the vehicle driving trace and the lane to improve and extend our driving behavior dataset.
\begin{figure}[!ht]
\centering
\subfloat[]{\begin{minipage}[]{\linewidth}
\includegraphics[width=\linewidth]{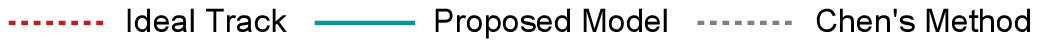}
\vfill
\vspace{0.1in}
\includegraphics[width=\linewidth]{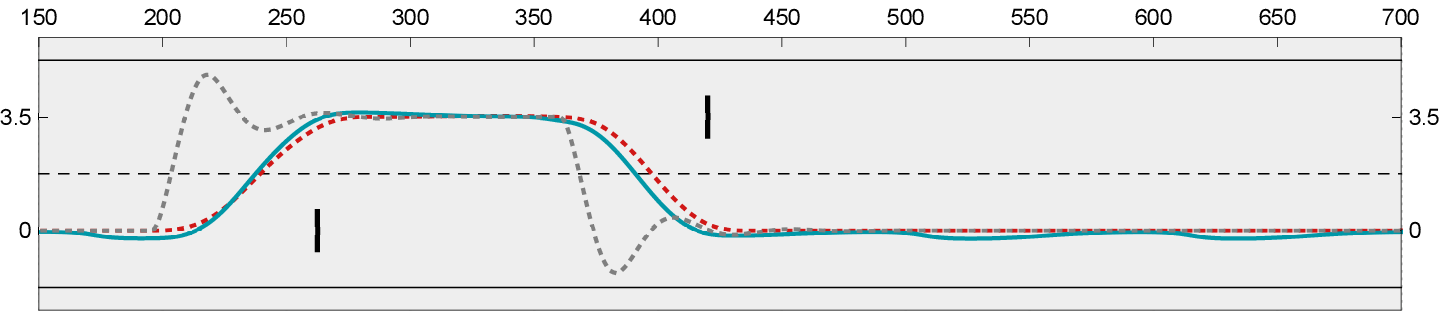}
\end{minipage}
\label{tracka}}

\subfloat[]{\includegraphics[width=3.5in]{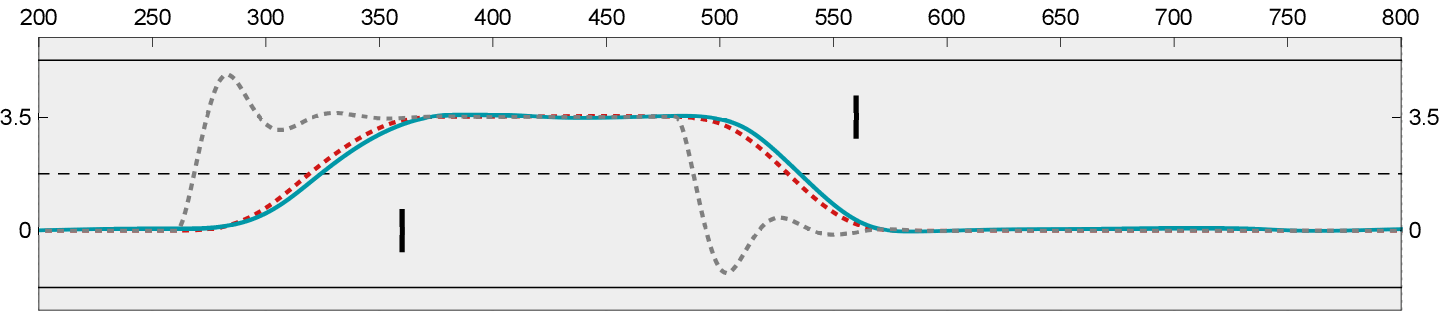}
\label{trackb}}

\subfloat[]{\includegraphics[width=3.5in]{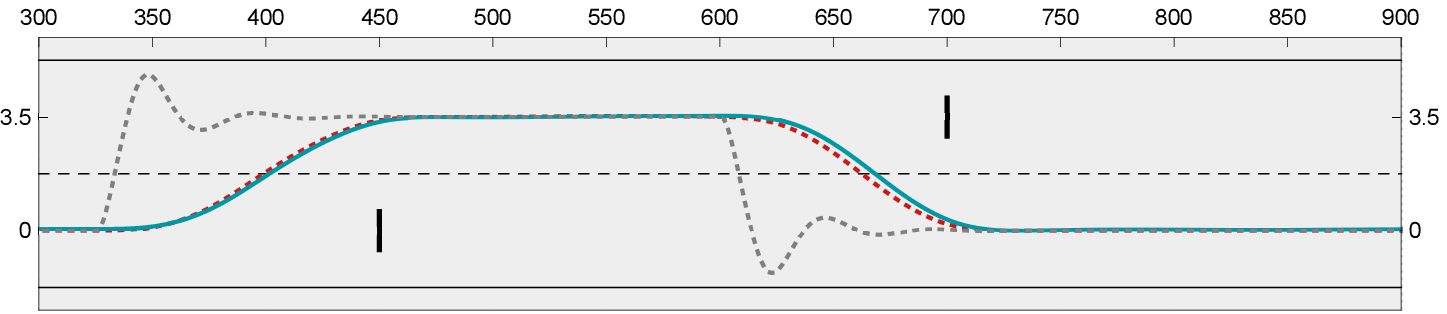}
\label{trackc}}

\subfloat[]{\includegraphics[width=3.5in]{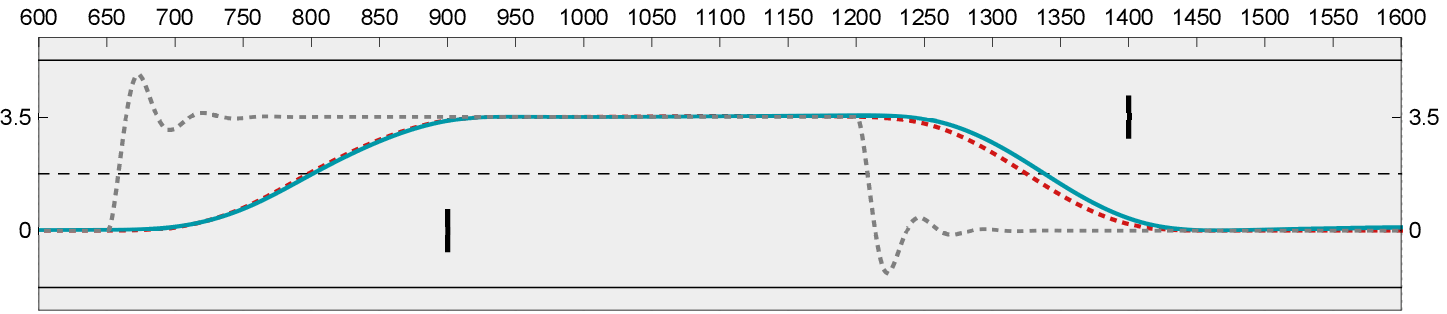}
\label{trackd}}

\subfloat[]{\includegraphics[width=3.5in]{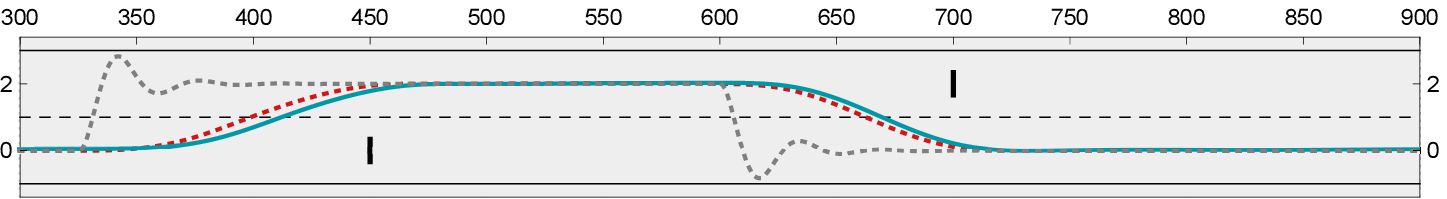}
\label{tracke}}

\subfloat[]{\includegraphics[width=3.5in]{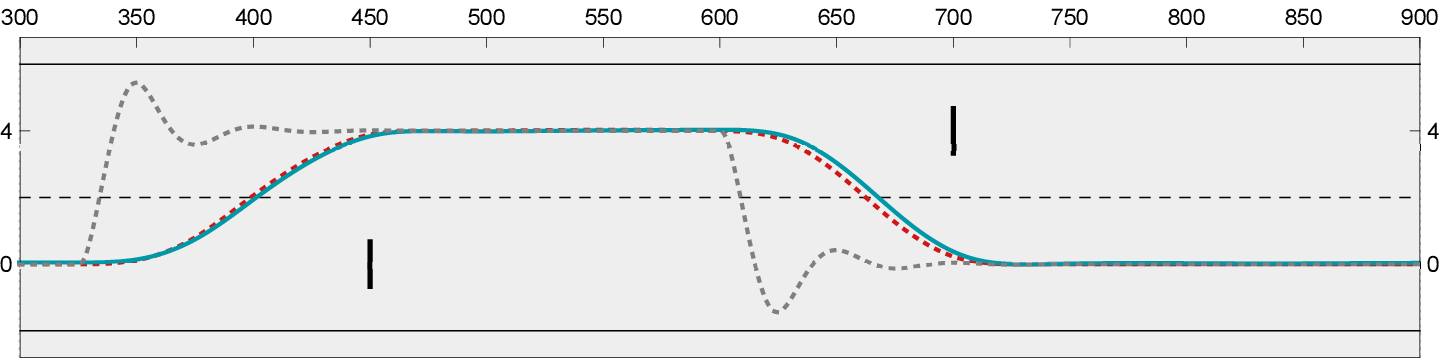}
\label{trackf}}

\caption{Simulation of planning and control process with CMA model.
Simulative vehicle trajectories on diverse lanes. From \ref{tracka} to \ref{trackf}, obstacles are put respectively at 50, 80, 100, 200, 100, 100 meters from the self-driving car, and occurs at random moments to test the planning and control ability of models. In \ref{tracka} - \ref{trackd}, the lanes are 3.5 meters in widths, and in \ref{tracke} - \ref{trackf}, the widths are of 4 meters.
}
\label{figure_sim_result}
\end{figure}

\begin{table}[!htb]
\renewcommand{\arraystretch}{1.2}
\caption{Parameters of recurrent neural network with lstm block}
\label{tablernn}
\centering
\begin{tabular}{c||rl}
\hline
\makecell[c]{\bfseries Layers} &
\makecell[c]{\bfseries Operations} & 
\makecell[c]{\bfseries Attributions} \\
% \hhline{=#==}
\hline\hline
\multirow{2}{*}{Dense}
& Input & Size:  $\left[20\times C^t \right]$\\
& Output & Size: $\left[20\times16\right]$ \\
\hline
\multirow{2}{*}{LSTM}
& Input & Size: $\left[20 \times 16\right]$ \\
& Output & Size: $\left[20 \times 64\right]$ \\
\hline
\multirow{2}{*}{LSTM}
& Input & Size: $\left[20 \times 64\right]$ \\
& Output & Size: $\left[20 \times 64\right]$ \\
\hline
\multirow{2}{*}{LSTM}
& Input & Size: $\left[20 \times 64\right]$ \\
& Output & Size: $\left[64\right]$ \\
\hline
\multirow{2}{*}{Dense}
& Input & Size: $\left[64\right]$ \\
& Output & Steering angle $S_a$ \\
\hline
\end{tabular}
\end{table}

%分析是否达到了human-level control
As shown in Table. \ref{tablernn}, we build a LSTM network to model the temporal dependencies in driving process. Path planning and vehicle control will be accomplished simultaneously in the LSTM network. In order to quantitative analyze the {efficiency} of our method in path planning and control, we evaluate the proposed method in the simulator, instead of implementing it on a real vehicle.
In the simulator, we set two lanes in the road, and the speed of vehicle is {set but not limited} to 40 Km/h. In the meanwhile, there will be two obstacles set on the lane to test if our model can achieve the control of human-level in the lane changing scenario. As shown in Fig. \ref{simulator}, the proposed method replaces the internal controller (driver model) in the simulator. 
A steering angle to control the car module in real-time can be generated from the proposed model by using information of car states and environment data.
% Car states and environment data will be input in our model, and it will output a steering angle to control the car module in real-time.

% As shown in Fig. \ref{figure_sim_result}, 
%我们呈现了，在不同的lane changing场景中，我们的方法与 chen‘s的方法的轨迹图。在汽车行驶的测试过程中，我们随机在车辆的前方加入障碍物，检验模型在面对不同距离的障碍物时所产生的换道操作的性能。我们的方法在换道操作中，与chen‘s的方法相比，由于考虑了时间上的依赖关系，记忆了车辆过去一段时间内的状态，所以行驶的轨迹与理想的曲线十分吻合，可以控制车辆平稳、平滑的行驶，整个过程与人类驾驶员的操作十分相似，充分学习到了人类驾驶员的行为。
%
As shown in Fig. \ref{figure_sim_result}, we present the driving trajectories generated by our method and Chen's method \cite{chen2015deepdriving} in different lane changing scenarios. In testing procedure, we randomly set obstructions in front of the vehicle to test its performance on lane changing operation when faced with obstacles of different distances. As for the lane changing operation, compared with Chen's method, our approach takes account of the temporal dependence, which implies the states of the vehicle over a period of time are memorized. Therefore, the trajectory of our model is consistent with the ideal curve, and the vehicle can drives in a steady and smooth state. The whole process is similar to the operation of human drivers, where the drivers' behaviors are absolutely learn by model.

% \subsubsection{Implementation Details} 
% In this part
% \subsubsection{The Effects of Control Sequence Generation}
% In this part

\section{conclusion}
In this paper, we proposed a cognitive model with attention for self-driving cars. 
% The proposed model is fully inspired by human brain, simulating human visual cortex and motor cortex in sensing, planning and control, while the mechanism of attention in brain is molded by recurrent neural network in time. To implement our model, the concept of cognitive map for traffic scene is defined and introduced in detail. 
This model was proposed, inspired by human brain, to simulate human visual and motor cortices for sensing, planning and control. The mechanism of attention was modeled by a recurrent neural network in time. In addition, the concept of cognitive map for traffic scene was introduced and described in detail. Furthermore, a labeled dataset named Road-Vehicle Dataset (RVD) is built for training and evaluating. 
% We evaluate the performance of our model in three visual tasks, and simulate the ability of our method in planning and control. The experimental results demonstrate that our proposed cognitive model is able to complete some basic self-driving tasks simply with cameras.
% We estimate the ability of our model by three visual tasks, and evaluate its performance in planning and control. The experimental results demonstrate that the proposed cognitive model is able to complete some basic self-driving tasks simply with cameras.
The performance of the proposed model in planning and control was tested by three visual tasks. Experimental results showed that our model can fulfill some basic self-driving tasks with only cameras.

Besides attention mechanism, the permanent memory plays a crucial role in human cognition. How to incorporate the permanent memory into the proposed cognitive model is left as our future work. In addition, there are many abnormal events in the actual traffic scene. How to develop an efficient cognitive model to deal with these situations is an interesting topic for future study.   
\section*{Acknowledgment}

% National Natural Science Foundation of China
This research was partially supported by the National Natural Science Foundation of China (No. 91520000, L1522023), the Programme of Introducing Talents of Discipline to University (No. B13043), 973 Program (No. 2015CB351703).

% , the National Natural Science Foundation of China (No. 61372152), Nation Key Research and Development Plan (No. 2016YFB0200202) and National High Technology Research and Development Program of China (No. 2014AA01A301).

\ifCLASSOPTIONcaptionsoff
  \newpage
\fi



%\bibliographystyle{IEEEtran}
%\begin{thebibliography}{1}
%\end{thebibliography}

\bibliographystyle{IEEEtran}
\bibliography{ref}

\begin{thebibliography}{10}
\providecommand{\url}[1]{#1}
\csname url@samestyle\endcsname
\providecommand{\newblock}{\relax}
\providecommand{\bibinfo}[2]{#2}
\providecommand{\BIBentrySTDinterwordspacing}{\spaceskip=0pt\relax}
\providecommand{\BIBentryALTinterwordstretchfactor}{4}
\providecommand{\BIBentryALTinterwordspacing}{\spaceskip=\fontdimen2\font plus
\BIBentryALTinterwordstretchfactor\fontdimen3\font minus
  \fontdimen4\font\relax}
\providecommand{\BIBforeignlanguage}[2]{{%
\expandafter\ifx\csname l@#1\endcsname\relax
\typeout{** WARNING: IEEEtran.bst: No hyphenation pattern has been}%
\typeout{** loaded for the language `#1'. Using the pattern for}%
\typeout{** the default language instead.}%
\else
\language=\csname l@#1\endcsname
\fi
#2}}
\providecommand{\BIBdecl}{\relax}
\BIBdecl

\bibitem{xue2017vision}
J.-r. Xue, D.~Wang, S.-y. Du, D.-x. Cui, Y.~Huang, and N.-n. Zheng, ``A
  vision-centered multi-sensor fusing approach to self-localization and
  obstacle perception for robotic cars,'' \emph{Front. Inform. Technol.
  Electron. Eng}, vol.~18, no.~1, pp. 122--138, 2017.

\bibitem{tsugawa1994vision}
S.~Tsugawa, ``Vision-based vehicles in japan: Machine vision systems and
  driving control systems,'' \emph{IEEE Transactions on Industrial
  Electronics}, vol.~41, no.~4, pp. 398--405, 1994.

\bibitem{turk1988vits}
M.~A. Turk, D.~G. Morgenthaler, K.~D. Gremban, and M.~Marra, ``Vits-a vision
  system for autonomous land vehicle navigation,'' \emph{IEEE Transactions on
  Pattern Analysis and Machine Intelligence}, vol.~10, no.~3, pp. 342--361,
  1988.

\bibitem{leonard2008perception}
J.~Leonard, J.~How, S.~Teller, M.~Berger, S.~Campbell, G.~Fiore, L.~Fletcher,
  E.~Frazzoli, A.~Huang, S.~Karaman \emph{et~al.}, ``A perception-driven
  autonomous urban vehicle,'' \emph{Journal of Field Robotics}, vol.~25,
  no.~10, pp. 727--774, 2008.

\bibitem{bojarski2016end}
M.~Bojarski, D.~Del~Testa, D.~Dworakowski, B.~Firner, B.~Flepp, P.~Goyal, L.~D.
  Jackel, M.~Monfort, U.~Muller, J.~Zhang \emph{et~al.}, ``End to end learning
  for self-driving cars,'' \emph{arXiv preprint arXiv:1604.07316}, 2016.

\bibitem{krizhevsky2012imagenet}
A.~Krizhevsky, I.~Sutskever, and G.~E. Hinton, ``Imagenet classification with
  deep convolutional neural networks,'' in \emph{Advances in neural information
  processing systems}, 2012, pp. 1097--1105.

\bibitem{lecun2015deep}
Y.~LeCun, Y.~Bengio, and G.~Hinton, ``Deep learning,'' \emph{Nature}, vol. 521,
  no. 7553, pp. 436--444, 2015.

\bibitem{Zheng2017hunhe}
N.~Zheng, Z.~Liu, P.~Ren, Y.~Ma, S.~Chen, S.~Yu, J.~Xue, B.~Chen, and F.~Wang,
  ``Hybrid-augmented intelligence: collaboration and cognition,'' vol.~18,
  no.~2, pp. 153--179, 2017.

\bibitem{goodfellow2016deep}
I.~Goodfellow, Y.~Bengio, and A.~Courville, \emph{Deep learning}.\hskip 1em
  plus 0.5em minus 0.4em\relax MIT Press, 2016.

\bibitem{hawkins2007intelligence}
J.~Hawkins and S.~Blakeslee, \emph{On intelligence}.\hskip 1em plus 0.5em minus
  0.4em\relax Macmillan, 2007.

\bibitem{Mountcastle1978}
V.~Mountcastle, ``{An organizing principle for cerebral function: the unit
  model and the distributed system},'' in \emph{The Mindful Brain}, G.~Edelman
  and V.~Mountcastle, Eds.\hskip 1em plus 0.5em minus 0.4em\relax Cambridge,
  Mass.: MIT Press, 1978.

\bibitem{tolman1948cognitive}
E.~C. Tolman \emph{et~al.}, ``Cognitive maps in rats and men,'' 1948.

\bibitem{mcnaughton2006path}
B.~L. McNaughton, F.~P. Battaglia, O.~Jensen, E.~I. Moser, and M.-B. Moser,
  ``Path integration and the neural basis of the'cognitive map','' \emph{Nature
  Reviews Neuroscience}, vol.~7, no.~8, pp. 663--678, 2006.

\bibitem{chen2015deepdriving}
C.~Chen, A.~Seff, A.~Kornhauser, and J.~Xiao, ``Deepdriving: Learning
  affordance for direct perception in autonomous driving,'' in
  \emph{Proceedings of the IEEE International Conference on Computer Vision},
  2015, pp. 2722--2730.

\bibitem{wang2004lane}
Y.~Wang, E.~K. Teoh, and D.~Shen, ``Lane detection and tracking using
  b-snake,'' \emph{Image and Vision computing}, vol.~22, no.~4, pp. 269--280,
  2004.

\bibitem{assidiq2008real}
A.~A. Assidiq, O.~O. Khalifa, M.~R. Islam, and S.~Khan, ``Real time lane
  detection for autonomous vehicles,'' in \emph{Computer and Communication
  Engineering, 2008. ICCCE 2008. International Conference on}.\hskip 1em plus
  0.5em minus 0.4em\relax IEEE, 2008, pp. 82--88.

\bibitem{li2014multiple}
Y.~Li, A.~Iqbal, and N.~R. Gans, ``Multiple lane boundary detection using a
  combination of low-level image features,'' in \emph{Intelligent
  Transportation Systems (ITSC), 2014 IEEE 17th International Conference
  on}.\hskip 1em plus 0.5em minus 0.4em\relax IEEE, 2014, pp. 1682--1687.

\bibitem{he2010lane}
J.~He, H.~Rong, J.~Gong, and W.~Huang, ``A lane detection method for lane
  departure warning system,'' in \emph{Optoelectronics and Image Processing
  (ICOIP), 2010 International Conference on}, vol.~1.\hskip 1em plus 0.5em
  minus 0.4em\relax IEEE, 2010, pp. 28--31.

\bibitem{nan2016efficient}
Z.~Nan, P.~Wei, L.~Xu, and N.~Zheng, ``Efficient lane boundary detection with
  spatial-temporal knowledge filtering,'' \emph{Sensors}, vol.~16, no.~8, p.
  1276, 2016.

\bibitem{huval2015empirical}
B.~Huval, T.~Wang, S.~Tandon, J.~Kiske, W.~Song, J.~Pazhayampallil,
  M.~Andriluka, P.~Rajpurkar, T.~Migimatsu, R.~Cheng-Yue \emph{et~al.}, ``An
  empirical evaluation of deep learning on highway driving,'' \emph{arXiv
  preprint arXiv:1504.01716}, 2015.

\bibitem{girshick2016region}
R.~Girshick, J.~Donahue, T.~Darrell, and J.~Malik, ``Region-based convolutional
  networks for accurate object detection and segmentation,'' \emph{IEEE
  transactions on pattern analysis and machine intelligence}, vol.~38, no.~1,
  pp. 142--158, 2016.

\bibitem{redmon2016you}
J.~Redmon, S.~Divvala, R.~Girshick, and A.~Farhadi, ``You only look once:
  Unified, real-time object detection,'' in \emph{Proceedings of the IEEE
  Conference on Computer Vision and Pattern Recognition}, 2016, pp. 779--788.

\bibitem{pomerleau1989alvinn}
D.~A. Pomerleau, ``Alvinn, an autonomous land vehicle in a neural network,''
  Carnegie Mellon University, Computer Science Department, Tech. Rep., 1989.

\bibitem{lecun2005off}
Y.~LeCun, U.~Muller, J.~Ben, E.~Cosatto, and B.~Flepp, ``Off-road obstacle
  avoidance through end-to-end learning,'' in \emph{NIPS}, 2005, pp. 739--746.

\bibitem{xu2016end}
H.~Xu, Y.~Gao, F.~Yu, and T.~Darrell, ``End-to-end learning of driving models
  from large-scale video datasets,'' \emph{arXiv preprint arXiv:1612.01079},
  2016.

\bibitem{koutnik2013evolving}
J.~Koutn{\'\i}k, G.~Cuccu, J.~Schmidhuber, and F.~Gomez, ``Evolving large-scale
  neural networks for vision-based torcs,'' 2013.

\bibitem{koutnik2013evolv}
------, ``Evolving large-scale neural networks for vision-based reinforcement
  learning,'' in \emph{Proceedings of the 15th annual conference on Genetic and
  evolutionary computation}.\hskip 1em plus 0.5em minus 0.4em\relax ACM, 2013,
  pp. 1061--1068.

\bibitem{sutton1998reinforcement}
R.~S. Sutton and A.~G. Barto, \emph{Reinforcement learning: An
  introduction}.\hskip 1em plus 0.5em minus 0.4em\relax MIT press Cambridge,
  1998, vol.~1, no.~1.

\bibitem{mnih2015human}
V.~Mnih, K.~Kavukcuoglu, D.~Silver, A.~A. Rusu, J.~Veness, M.~G. Bellemare,
  A.~Graves, M.~Riedmiller, A.~K. Fidjeland, G.~Ostrovski \emph{et~al.},
  ``Human-level control through deep reinforcement learning,'' \emph{Nature},
  vol. 518, no. 7540, pp. 529--533, 2015.

\bibitem{silver2016mastering}
D.~Silver, A.~Huang, C.~J. Maddison, A.~Guez, L.~Sifre, G.~Van Den~Driessche,
  J.~Schrittwieser, I.~Antonoglou, V.~Panneershelvam, M.~Lanctot \emph{et~al.},
  ``Mastering the game of go with deep neural networks and tree search,''
  \emph{Nature}, vol. 529, no. 7587, pp. 484--489, 2016.

\bibitem{mnih2013playing}
V.~Mnih, K.~Kavukcuoglu, D.~Silver, A.~Graves, I.~Antonoglou, D.~Wierstra, and
  M.~Riedmiller, ``Playing atari with deep reinforcement learning,''
  \emph{arXiv preprint arXiv:1312.5602}, 2013.

\bibitem{paden2016survey}
B.~Paden, M.~{\v{C}}{\'a}p, S.~Z. Yong, D.~Yershov, and E.~Frazzoli, ``A survey
  of motion planning and control techniques for self-driving urban vehicles,''
  \emph{IEEE Transactions on Intelligent Vehicles}, vol.~1, no.~1, pp. 33--55,
  2016.

\bibitem{li2004calibration}
Q.~Li, N.-n. Zheng, and X.-t. Zhang, ``Calibration of external parameters of
  vehicle-mounted camera with trilinear method,'' \emph{Opto-electronic
  Engineering}, vol.~31, no.~8, p.~23, 2004.

\bibitem{zhang2000flexible}
Z.~Zhang, ``A flexible new technique for camera calibration,'' \emph{IEEE
  Transactions on pattern analysis and machine intelligence}, vol.~22, no.~11,
  pp. 1330--1334, 2000.

\bibitem{abadi2016tensorflow}
M.~Abadi, P.~Barham, J.~Chen, Z.~Chen, A.~Davis, J.~Dean, M.~Devin,
  S.~Ghemawat, G.~Irving, M.~Isard \emph{et~al.}, ``Tensorflow: A system for
  large-scale machine learning,'' in \emph{Proceedings of the 12th USENIX
  Symposium on Operating Systems Design and Implementation (OSDI). Savannah,
  Georgia, USA}, 2016.

\bibitem{fritsch2013new}
J.~Fritsch, T.~Kuhnl, and A.~Geiger, ``A new performance measure and evaluation
  benchmark for road detection algorithms,'' in \emph{Intelligent
  Transportation Systems-(ITSC), 2013 16th International IEEE Conference
  on}.\hskip 1em plus 0.5em minus 0.4em\relax IEEE, 2013, pp. 1693--1700.

\end{thebibliography}


%\end{IEEEbiography}
%% insert where needed to balance the two columns on the last page with
%%\newpage


%\vfill


\end{document}